\documentclass[10pt,twocolumn,letterpaper]{article}

\usepackage[pagenumbers]{cvpr}

\usepackage{fancyhdr}
\usepackage{tikz}
\usepackage{siunitx}
\usepackage{tabularx}
\usepackage{pifont}
\usepackage{stfloats}
\usepackage{multirow}
\usepackage[table]{xcolor}
\usepackage{colortbl}

\usepackage{graphicx}
\usepackage{subcaption}
\usepackage{adjustbox}

\definecolor{citationblue}{RGB}{0, 113, 188}
\newcommand{\cmark}{\ding{51}}
\newcommand{\xmark}{\ding{55}}

\newcommand{\rurl}[1]{\href{http://#1}{\nolinkurl{#1}}}
\newcolumntype{d}[1]{S[table-format=#1, table-align-text-post=false]}
\definecolor{SoftGreen}{HTML}{D4EDDA}
\definecolor{SoftOrange}{HTML}{FFD8B1}
\definecolor{SoftRed}{HTML}{FFB3B3}

\makeatletter
\def\ps@firstpage{%
  \let\@oddhead\@empty
  \let\@evenhead\@empty
  \def\@oddfoot{\hfil\small\textit{IEEE/CVF Conference on Computer Vision and Pattern Recognition (CVPR 2026)}\hfil}%
  \let\@evenfoot\@oddfoot
}
\makeatother

\definecolor{cvprblue}{rgb}{0.21,0.49,0.74}
\usepackage[pagebackref,breaklinks,colorlinks,allcolors=cvprblue]{hyperref}

\title{TerraSeg: Self-Supervised Ground Segmentation for Any LiDAR}

\author{
Ted Lentsch$^{1}$ \quad Santiago Montiel-Marín$^{2}$ \quad Holger Caesar$^{1}$ \quad Dariu M. Gavrila$^{1}$ \\[2pt]
$^{1}$Department of Cognitive Robotics, Delft University of Technology \\
$^{2}$Department of Electronics, University of Alcalá
}

\begin{document}

\twocolumn[{%
\maketitle
\vspace{-9mm}
\begin{center}
    \includegraphics[width=0.98\textwidth]{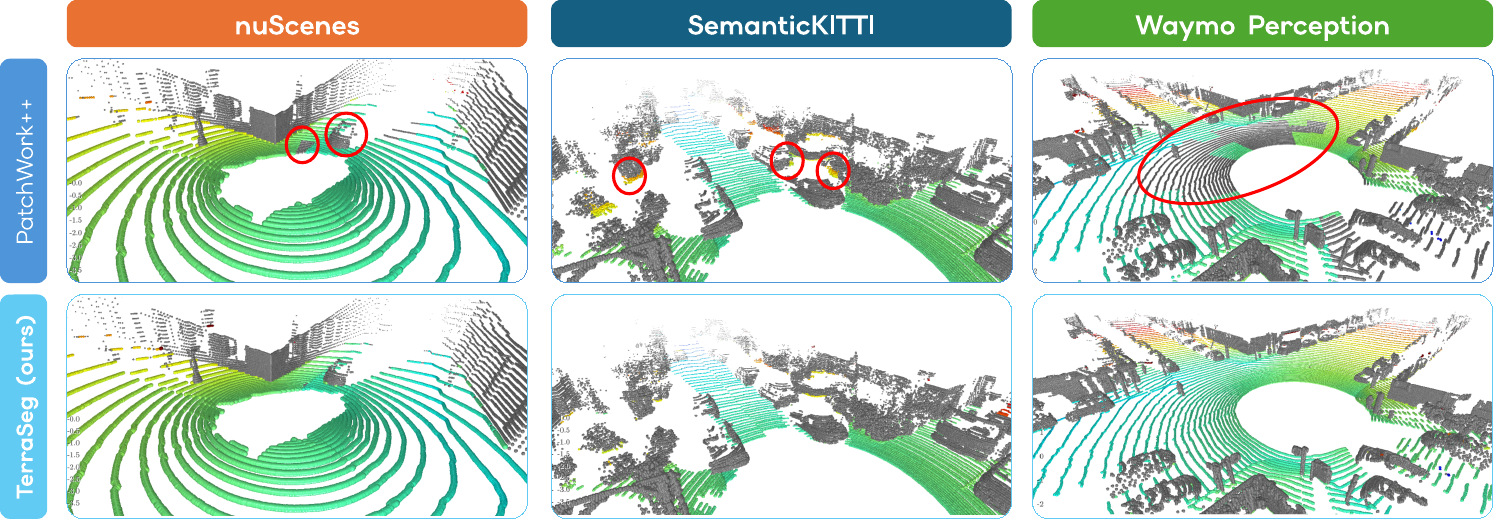}
    \captionof{figure}{
    \textbf{Ground segmentation performance across diverse LiDAR benchmarks for baseline PatchWork++~\cite{lee2022patchwork++} and our \emph{TerraSeg}.}
    We present a qualitative comparison between TerraSeg and the baseline on the nuScenes~\cite{caesar2020nuscenes}, SemanticKITTI~\cite{behley2019semantickitti}, and Waymo Perception~\cite{sun2020scalability} datasets.
    Ground points are colored based on elevation, non-ground points are gray, and highlighted regions are marked with red.
    }\label{fig:teaser}
    \vspace{1mm}
\end{center}%
}]

\thispagestyle{firstpage}

\begin{abstract}
    LiDAR perception is fundamental to robotics, enabling machines to understand their environment in 3D. A crucial task for LiDAR-based scene understanding and navigation is ground segmentation. However, existing methods are either handcrafted for specific sensor configurations or rely on costly per-point manual labels, severely limiting their generalization and scalability. To overcome this, we introduce TerraSeg, the first self-supervised, domain-agnostic model for LiDAR ground segmentation. We train TerraSeg on OmniLiDAR, a unified large-scale dataset that aggregates and standardizes data from 12 major public benchmarks. Spanning almost 22 million raw scans across 15 distinct sensor models, OmniLiDAR provides unprecedented diversity for learning a highly generalizable ground model. To supervise training without human annotations, we propose PseudoLabeler, a novel module that generates high-quality ground and non-ground labels through self-supervised per-scan runtime optimization. Extensive evaluations demonstrate that, despite using no manual labels, TerraSeg achieves state-of-the-art results on nuScenes, SemanticKITTI, and Waymo Perception while delivering real-time performance.
    Our \href{https://www.github.com/TedLentsch/TerraSeg}{code} and \href{https://huggingface.co/TedLentsch/TerraSeg}{model weights} are publicly available.
\end{abstract}
\vspace{-4.9mm}
    
\section{Introduction}\label{sec:introduction}
\vspace{1mm}

\noindent Mobile robots~\cite{de2025vehicle, oliveira2021advances} are rapidly emerging in applications such as warehouse automation, robotaxis, and precision agriculture.
To navigate safely and interact effectively with complex, unstructured environments, these robots rely on LiDAR sensors to perceive the 3D geometry of their surroundings.
In particular, reliable ground segmentation is essential for downstream tasks such as object discovery, free space estimation, and localization and mapping.

A mobile robot's perception stack often combines two object detection pipelines: 
(1) an \emph{explicit} pipeline, where data-driven approaches learn to detect objects belonging to a predefined set of object classes, and 
(2) an \emph{implicit} pipeline, where rule-based steps such as ground segmentation followed by spatial clustering yield object proposals for arbitrary classes.
For safety, the implicit pipeline frequently serves as a fallback when the explicit one fails, making fast and accurate ground segmentation critical for \emph{online} operation.
Ground segmentation is equally vital in \emph{offline} label-efficient workflows such as unsupervised or weakly supervised auto-labeling for 3D object detection~\cite{lentsch2024union, khurana2025shelf} and self-supervised 3D representation learning~\cite{nunes2023temporal, wu2023spatiotemporal}.
Errors in ground modeling propagate to downstream tasks, \eg overestimating the ground inflates free-space.
Furthermore, ground segmentation serves as an effective lightweight pre-processing primitive that separates the ground from foreground regions of interest, significantly improving the efficiency of downstream perception pipelines.

Despite its importance, handcrafted methods~\cite{lee2022patchwork++, oh2022travel} are typically tuned on single datasets and require re-tuning for new environments, varying weather conditions, or sensor configurations. 
While geometric methods like RANSAC~\cite{fischler1981random} are fast, their global planar assumptions prove brittle on non-planar terrain, highlighting the need for flexible, locally adaptive priors. 
Conversely, supervised data-driven methods~\cite{paigwar2020gndnet} generalize better but rely on expensive per-point manual annotations. 
This motivates label-free methods capable of zero-shot domain generalization.

\begin{figure*}[!t]
  \centering
  \includegraphics[width=0.995\textwidth]{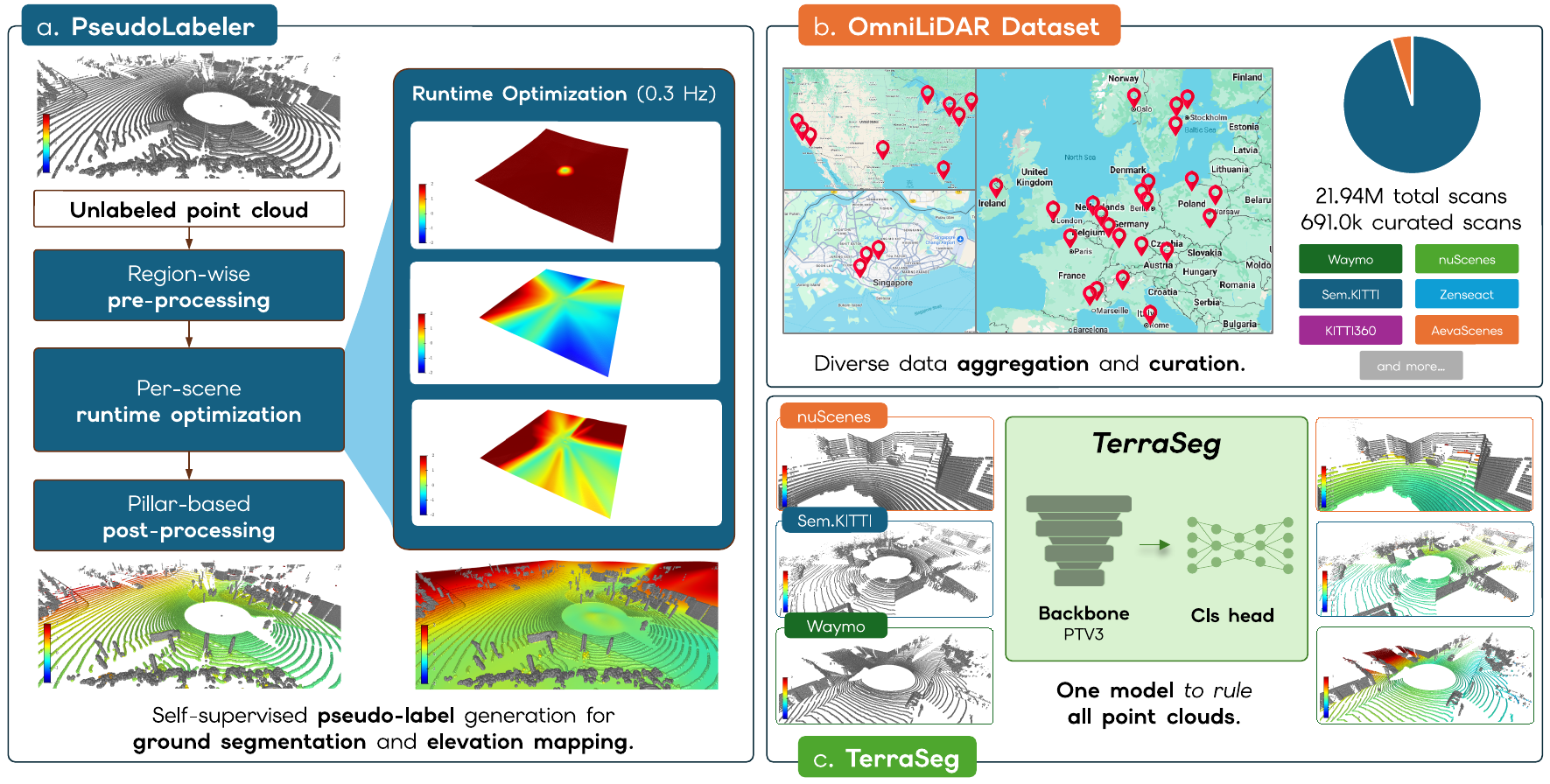}
  \caption{
    \textbf{Overview of TerraSeg.}
    (a) PseudoLabeler generates self-supervised point-wise ground/non-ground labels per raw LiDAR scan.
    (b) OmniLiDAR unifies 12 major automotive datasets within an aggregated framework, yielding a diverse curated corpus drawn from nearly 22 million raw scans.
    (c) TerraSeg is a real-time, domain-agnostic model for ground segmentation, trained on OmniLiDAR using these self-supervised pseudo-labels. 
    We introduce two architectures: accurate Base model (TerraSeg-B) and efficient Small model (TerraSeg-S).
  }
  \vspace{-2mm}
  \label{fig:overview}
\end{figure*}

Recently, large-scale pre-training has enabled robust generalization in Natural Language Processing~\cite{guo2025deepseek} and Computer Vision~\cite{simeoni2025dinov3, carion2025sam}, reducing the need for dataset-specific domain adaptation. 
However, its application for cross-sensor LiDAR generalization remains unexplored. 
Rather than pursuing a universal multi-task system, we argue for a single-task, domain-agnostic approach: training a self-supervised model on highly diverse geometry to achieve zero-shot transfer across datasets and sensors.

We present \emph{TerraSeg}, the first self-supervised, domain-agnostic model for LiDAR ground segmentation (\Cref{fig:teaser}). 
Our approach entails curating a massive dataset, generating high-quality pseudo-labels via self-supervised learning, and using these pseudo-labels to supervise a fast, deployable network. 
We introduce the \emph{OmniLiDAR} dataset, unifying \num{12} public driving datasets by standardizing raw scans and harmonizing labels for cross-dataset benchmarking. 
This yields a balanced, diverse corpus of nearly \num{22} million scans across \num{15} unique LiDARs. 
Each scan is pseudo-labeled by \emph{PseudoLabeler} via self-supervised runtime optimization to fit a global smooth elevation map. 
Finally, we train \emph{TerraSeg} on these labels using an adapted version of Point Transformer v3~\cite{wu2024point}. 
To support varied compute constraints, we offer an accurate Base model (TerraSeg-B) running at \qtyrange{10}{28}{\hertz}, and an efficient Small model (TerraSeg-S) at \qtyrange{17}{50}{\hertz}. 
The approach is summarized in~\Cref{fig:overview} and our key contributions are three-fold:
\begin{enumerate}
    \item \textbf{OmniLiDAR}: Unified outdoor LiDAR dataset merging scans from \num{12} public datasets, totaling a diverse corpus of nearly \num{22} million raw scans from \num{15} distinct sensors, with standardized geometry and harmonized labels.
    \item \textbf{PseudoLabeler}: Self-supervised module that generates high-quality ground pseudo-labels via per-scan runtime optimization, robust to noise and varying point densities.
    \item \textbf{TerraSeg}: First self-supervised, domain-agnostic model for LiDAR ground segmentation; providing real-time capabilities across scalable Base and Small architectures, and achieving new state-of-the-art on the nuScenes, SemanticKITTI, and Waymo Perception datasets.
\end{enumerate}

\section{Related work}\label{sec:related_work}

\begin{table*}[!t]
    \centering
    \caption{
    \textbf{Overview of main methods for ground segmentation.}
    In the \emph{Method type} column, \emph{Handcrafted} refers to rule-based, non-learnable algorithms, while \emph{Supervised} and \emph{Self-supervised} indicate learning-based approaches.
    In the \emph{Tasks} columns, \emph{Seg.} and \emph{Elev.} mean ground \emph{Segmentation} and \emph{Elevation mapping}, respectively.
    $^\dagger$Implemented in TorchRobotics~\cite{lentsch2025torchrobotics}.
    $^\ddagger$Confidence is bin-level only, used internally for classification, not output as a per-point score.
    $^\S$We obtained source code from the authors but the project is not open source.
    }
    \label{tab:baseline_comparison}
    \begin{tabularx}{\textwidth}{>{\raggedright\arraybackslash}p{3.1cm} | >{\centering\arraybackslash}p{0.9cm} | >{\centering\arraybackslash}p{2.6cm} | >{\centering\arraybackslash}p{0.8cm} >{\centering\arraybackslash}p{0.8cm} | >{\centering\arraybackslash}p{1.6cm} >{\centering\arraybackslash}p{1.5cm} | >{\centering\arraybackslash}p{1.2cm} >{\centering\arraybackslash}p{1.2cm}}
        \toprule
        \multicolumn{1}{c}{} & \multicolumn{1}{c}{} & \multicolumn{1}{c}{} & \multicolumn{2}{c}{Tasks} & \multicolumn{2}{c}{Attributes} &
        \multicolumn{2}{c}{Code release} \\
        \multicolumn{1}{l}{Method} & \multicolumn{1}{c}{Year} & \multicolumn{1}{c}{Method type} & \multicolumn{1}{c}{Seg.} & \multicolumn{1}{c}{Elev.} & \multicolumn{1}{c}{Confidence} & \multicolumn{1}{c}{Real-time} & \multicolumn{1}{c}{Python} & \multicolumn{1}{c}{ROS2} \\
        \midrule
        RANSAC~\cite{fischler1981random} & 1981 & Handcrafted & \textcolor{citationblue}{\cmark} & \textcolor{citationblue}{\cmark} & \textcolor{gray}{\xmark} & \textcolor{citationblue}{\cmark} & \phantom{$^\dagger$}N/A$^\dagger$ & N/A \\
        GndNet~\cite{paigwar2020gndnet} & 2020 & Supervised & \textcolor{citationblue}{\cmark} & \textcolor{citationblue}{\cmark} & \textcolor{gray}{\xmark} & \textcolor{citationblue}{\cmark} & \textcolor{citationblue}{\cmark} & \textcolor{gray}{\xmark} \\
        PatchWork~\cite{lim2021patchwork} & 2021 & Handcrafted & \textcolor{citationblue}{\cmark} & \textcolor{gray}{\xmark} & \phantom{$^\ddagger$}\textcolor{gray}{\xmark}$^\ddagger$ & \textcolor{citationblue}{\cmark} & \textcolor{gray}{\xmark} & \textcolor{citationblue}{\cmark} \\
        JCP~\cite{shen2021fast} & 2021 & Handcrafted & \textcolor{citationblue}{\cmark} & \textcolor{gray}{\xmark} & \textcolor{citationblue}{\cmark} & \textcolor{citationblue}{\cmark} & \textcolor{gray}{\xmark} & \textcolor{gray}{\xmark} \\
        PatchWork++~\cite{lee2022patchwork++} & 2022 & Handcrafted & \textcolor{citationblue}{\cmark} & \textcolor{gray}{\xmark} & \phantom{$^\ddagger$}\textcolor{gray}{\xmark}$^\ddagger$ & \textcolor{citationblue}{\cmark} & \textcolor{citationblue}{\cmark} & \textcolor{citationblue}{\cmark} \\
        TRAVEL~\cite{oh2022travel} & 2022 & Handcrafted & \textcolor{citationblue}{\cmark} & \textcolor{gray}{\xmark} & \textcolor{gray}{\xmark} & \textcolor{citationblue}{\cmark} & \textcolor{gray}{\xmark} & \textcolor{gray}{\xmark} \\
        GroundGrid~\cite{steinke2023groundgrid} & 2023 & Handcrafted & \textcolor{citationblue}{\cmark} & \textcolor{citationblue}{\cmark} & \textcolor{citationblue}{\cmark} & \textcolor{citationblue}{\cmark} & \textcolor{gray}{\xmark} & \textcolor{gray}{\xmark} \\
        B‑TMS~\cite{oh2024b} & 2024 & Handcrafted & \textcolor{citationblue}{\cmark} & \textcolor{gray}{\xmark} & \textcolor{gray}{\xmark} & \textcolor{citationblue}{\cmark} & \textcolor{gray}{\xmark} & \textcolor{gray}{\xmark} \\
        GATA~\cite{del2024probabilistic} & 2024 & Handcrafted & \textcolor{citationblue}{\cmark} & \textcolor{citationblue}{\cmark} & \textcolor{citationblue}{\cmark} & \textcolor{citationblue}{\cmark} & \textcolor{gray}{\xmark} & \textcolor{gray}{\xmark} \\
        Chodosh et al.~\cite{chodosh2024re} & 2024 & Self-supervised & \textcolor{citationblue}{\cmark} & \textcolor{citationblue}{\cmark} & \textcolor{gray}{\xmark} & \textcolor{gray}{\xmark} & \phantom{$^\S$}\textcolor{citationblue}{\cmark}$^\S$ & \textcolor{gray}{\xmark} \\
        \midrule
        PseudoLabeler (ours) & 2026 & Self-supervised & \textcolor{citationblue}{\cmark} & \textcolor{citationblue}{\cmark} & \textcolor{gray}{\xmark} & \textcolor{gray}{\xmark} & \textcolor{citationblue}{\cmark} & \textcolor{gray}{\xmark} \\
        TerraSeg (ours) & 2026 & Self-supervised & \textcolor{citationblue}{\cmark} & \textcolor{gray}{\xmark} & \textcolor{citationblue}{\cmark} & \textcolor{citationblue}{\cmark} & \textcolor{citationblue}{\cmark} & \textcolor{citationblue}{\cmark} \\
        \bottomrule
    \end{tabularx}
\end{table*}

\noindent Here we review the work related to LiDAR ground segmentation.
We summarize the main approaches in \Cref{tab:baseline_comparison}.

\textbf{Handcrafted geometric.}
Ground segmentation aims to assign point-wise ground or non-ground labels to an unstructured LiDAR point cloud. Classical methods rely on surface-fitting algorithms or geometric heuristics to segment point clouds without learning. RANSAC~\cite{fischler1981random} operates under the assumption that the plane containing the largest number of points within a distance threshold corresponds to the ground plane. By iteratively fitting random planes, the algorithm often converges to a plausible solution. But due to the diversity of outdoor scenarios, a single plane may underfit non-flat scenes.
In contrast, PatchWork~\cite{lim2021patchwork} partitions a LiDAR scan into multiple regions using a concentric-zone polar grid and performs region-wise PCA plane fitting. Points within each region are classified as ground or non-ground using a probabilistical likelihood test. However, mixed bins may be uniformly labeled, leading to misclassification of regions containing both ground and non-ground points.
PatchWork++~\cite{lee2022patchwork++} extends PatchWork by introducing adaptive parameter tuning, temporal consistency mechanisms, and enhanced robustness to noise and multi-layered ground surfaces.
JCP~\cite{shen2021fast} projects LiDAR points to a 2D range view and applies slope-based heuristics to obtain an elevation change map as input to a coarse-to-fine ground segmentation.
Lastly, TRAVEL~\cite{oh2022travel} approaches ground segmentation by representing LiDAR scans as a node-edge graph by means of a Tri-Grid Field (TGF) graph structure. Traversable ground is modeled by analyzing the concavity/convexity of edges that connect nodes and segments.

Elevation mapping based methods produce a dense height map by assigning each LiDAR point to a grid cell and classifying it as ground or non-ground based on an elevation threshold. GroundGrid~\cite{steinke2023groundgrid} projects points into an elevation grid and detects ground cells via local height variance, followed by interpolation to fill missing regions. B-TMS~\cite{oh2024b} models terrain height using a Bayesian generalized kernel over a TGF, while GATA~\cite{del2024probabilistic} performs real-time terrain estimation with a probabilistic graph-based formulation.

In summary, handcrafted methods are fast and label-free but rely on simple terrain assumptions and sensor-specific tuning, motivating learning-based alternatives.

\textbf{Supervised learning.}
Learning-based approaches address the limitations of handcrafted pipelines, such as sensor-specific tuning and simplified ground models, by training neural networks to infer ground regions directly. GndNet~\cite{paigwar2020gndnet} builds on PointPillars~\cite{lang2019pointpillars} with a convolutional encoder-decoder that predicts a dense elevation map, enabling real-time ground segmentation by thresholding above-ground points. Its main limitation is the reliance on elevation labels produced by a handcrafted morphological pipeline, which restricts scalability and confines spatial coverage to the local neighborhood around each point cloud.

Although general-purpose 3D segmentation backbones like MinkowskiNet~\cite{choy20194d}, KPConv~\cite{thomas2019kpconv}, RandLA-Net~\cite{hu2020randla}, Cylinder3D~\cite{zhu2021cylindrical}, and the Point Transformer family~\cite{zhao2021point, wu2022point, wu2024point} offer strong representational capabilities for complex 3D scenes, their supervised training paradigms rely heavily on costly per-point manual annotations, limiting their practical scalability for LiDAR ground segmentation.

\textbf{Self-supervised learning.}
Chodosh et al.~\cite{chodosh2024re} use self-supervised runtime optimization to fit a piecewise-linear surface over the $(x,y)$ plane using a one-sided robust loss, and classify points as ground or non-ground by thresholding their vertical distance to this surface.
Although runtime optimization is too slow for online applications, we note that it can be used to pseudo-label large amounts of raw data.
We build on top of the method from Chodosh et al. and create our PseudoLabeler for labeling our OmniLiDAR dataset.
Compared to Chodosh et al, our novel PseudoLabeler has improved pre- and post-processing for ground segmentation, improved loss objective, and significantly higher throughput.
We present this in more detail in~\Cref{sec:method}.

\section{Methodology}\label{sec:method}

\begin{table*}[!t]
    \centering
    \caption{
    \textbf{OmniLiDAR dataset.}
    We unify \num{12} major public autonomous driving datasets into a single standardized LiDAR format, enabling large-scale cross-dataset training and evaluation.
    We curate the LiDAR scans to obtain a diverse training set by downsampling the LiDAR sequences to \qty{0.2}{\hertz}, \ie one LiDAR scan every five seconds.
    During training, we mitigate dataset imbalance by sampling with dataset-specific probabilities (see Section~\ref{subsec:OmniLiDAR}).
    All reported statistics are computed over the available LiDAR data, as some datasets provide only partial sensor releases.
    $^\dagger$Two different sensor configurations were used for data collection; we report the total number of distinct LiDARs used across configurations.
    $^\ddagger$Only 3D LiDAR sensors are considered.
    $^\S$Across all datasets, we identify \num{15} unique LiDAR hardware models.
    }
    \label{tab:omnilidar_overview}
    \begin{tabularx}{\textwidth}{
    >{\raggedright\arraybackslash}p{4.2cm} |
    >{\centering\arraybackslash}p{0.8cm} |
    d{2.0} |
    >{\centering\arraybackslash}p{0.9cm} |
    d{3.2} |
    d{2.3} |
    d{2.3} |
    >{\centering\arraybackslash}p{1.8cm}
}
        \toprule
        \multicolumn{1}{l}{Dataset} & \multicolumn{1}{c}{Year} & \multicolumn{1}{c}{\# LiDARs} & \multicolumn{1}{c}{Labels} & \multicolumn{1}{c}{Duration} & \multicolumn{1}{c}{\# Raw scans} & \multicolumn{1}{c}{\# Train scans} & \multicolumn{1}{c}{\# Eval scans} \\
        \midrule
        SemanticKITTI~\cite{behley2019semantickitti} & 2019 & 1 & \textcolor{citationblue}{\cmark} & 1.3\ h & 0.04\ M & 0.4\ k & 4.1\ k \\
        Lyft Level 5~\cite{kesten2019lyft} & 2019 & 4$^\dagger$ & \textcolor{gray}{\xmark} & 2.8\ h & 0.05\ M & 2.3\ k & - \\
        nuScenes~\cite{caesar2020nuscenes} & 2020 & 1 & \textcolor{citationblue}{\cmark} & 5.4\ h & 0.39\ M & 3.0\ k & 6.0\ k \\
        Waymo Perception~\cite{sun2020scalability} & 2020 & 5 & \textcolor{citationblue}{\cmark} & 6.3\ h & 1.14\ M & 15.1\ k & 6.0\ k \\
        PandaSet~\cite{xiao2021pandaset} & 2020 & 2 & \textcolor{citationblue}{\cmark} & 0.2\ h & 0.02\ M & 0.4\ k & - \\
        ONCE~\cite{mao2021one} & 2021 & 1 & \textcolor{gray}{\xmark} & 158.2\ h & 1.02\ M & 101.1\ k & - \\
        KITTI-360~\cite{liao2022kitti} & 2022 & 1$^\ddagger$ & \textcolor{gray}{\xmark} & 2.6\ h & 0.09\ M & 1.8\ k & - \\
        View-of-Delft~\cite{palffy2022multi} & 2022 & 1 & \textcolor{gray}{\xmark} & 0.2\ h & 0.01\ M & 0.2\ k & - \\
        Argoverse 2 Lidar~\cite{wilson2023argoverse} & 2023 & 2 & \textcolor{gray}{\xmark} & 165.4\ h & 11.95\ M & 236.9\ k & - \\
        Zenseact Open Dataset~\cite{alibeigi2023zenseact} & 2023 & 3 & \textcolor{gray}{\xmark} & 55.6\ h & 6.30\ M & 300.0\ k & - \\
        TruckScenes~\cite{fent2024man} & 2024 & 6 & \textcolor{gray}{\xmark} & 4.0\ h & 0.88\ M & 12.6\ k & - \\
        AevaScenes~\cite{narasimhan2025aevascenes} & 2025 & 6 & \textcolor{citationblue}{\cmark} & 0.3\ h & 0.06\ M & 1.2\ k & - \\
        \midrule
        \noalign{\global\arrayrulewidth=0pt}
        Total & & 33$^\S$ & & 402.3\ h & 21.94\ M & 675.0\ k & 16.1\ k \\
        \noalign{\global\arrayrulewidth=.4pt}
        \bottomrule
    \end{tabularx}
\end{table*}

We introduce LiDAR ground segmentation, detail our \emph{OmniLiDAR} dataset and self-supervised \emph{PseudoLabeler}, and present \emph{TerraSeg}, our domain-agnostic model for this task.

\subsection{LiDAR ground segmentation}\label{subsec:tasks}

In general, LiDAR segmentation aims to classify each point as one of the predefined semantic classes, such as \emph{road} and \emph{building}.
For ground segmentation, we generalize this to the binary case, \ie the \emph{ground} and \emph{non-ground} classes.
We develop a framework for ground segmentation without relying on manual labels that is compatible with all LiDARs.
We assume multiple mobile robots, each equipped with at least one time-synchronized LiDAR, without relying on any additional sensors (\eg camera, GPS/GNSS, IMU) to maximize usable data.
By aggregating LiDAR data from multiple robots, diverse geographic environments, and different sensor types, we achieve substantial data diversity.
This diversity enables our self-supervised, domain-agnostic model to achieve zero-shot ground segmentation across varying sensors. 
We define the inputs and outputs as follows:

\textbf{Input.}
The input to our method are the raw LiDAR scans collected during the traversals of the robots.
We assume that a traversal consists of a sequence of $T$ time steps, with a single LiDAR point cloud available at each time step $t$.
Let $P_t \in \mathbb{R}^{L \times 3}$ represent the 3D point cloud containing $L$ points. 
Although data is collected sequentially, our method processes each point cloud $P_t$ independently at the per-frame level, without relying on temporal fusion.

\textbf{Output.}
Our method outputs a set of predicted per-point confidence scores for binary classification.
Formally, the model maps the input point cloud into a continuous probability space, yielding a predicted score vector $\hat{y} \in [0,1]^{L}$.
Each element $\hat{y}_{i}$ represents the predicted probability that the $i$-th LiDAR point belongs to the \emph{non-ground} class.
To obtain the final discrete segmentation mask for evaluation or downstream autonomous driving tasks, these continuous probabilities are binarized using a tuned decision threshold.

\subsection{OmniLiDAR dataset}\label{subsec:OmniLiDAR}

We construct OmniLiDAR by aggregating nearly \num{22} million raw LiDAR sweeps drawn from \num{12} diverse and relevant autonomous driving benchmarks into a single, consistent format (see Table~\ref{tab:omnilidar_overview}).
Spanning \num{15} distinct LiDAR hardware models, this corpus provides unprecedented diversity for learning a highly generalizable ground segmentation model.

To ensure this dataset supports robust cross-sensor generalization, we apply a rigorous standardization process.
First, we curate a diverse training set by downsampling the LiDAR sequences to \qty{0.2}{Hz}.
Second, we align the LiDAR coordinate systems via sensor-specific transforms so that the horizontal $z=0$ plane is approximately ground-aligned and the positive $x$-axis points in the robot's forward moving direction.
Third, we apply a unique ego-vehicle removal radius tailored to each specific LiDAR sensor.
Each sweep retains only its standardized $(x,y,z)$ coordinates and metadata tags (identifying the dataset of origin, subset split, and sequence ID).
We treat each scan independently; no inter-sweep registration or multi-frame accumulation is performed.
Relying solely on standardized 3D coordinates ensures a lightweight, sensor-agnostic pipeline. Finally, we harmonize ground labels across datasets~\cite{caesar2020nuscenes, behley2019semantickitti, sun2020scalability, narasimhan2025aevascenes, xiao2021pandaset} to enable unified LiDAR ground segmentation evaluation.

We omit official test splits to prevent downstream data leakage.
To mitigate dataset imbalance and avoid bias toward massive collections, such as Argoverse~2 Lidar, we sample the source datasets during training using assigned probability weights.
See the Supplementary Material for the exact sampling distributions and implementation details.

\subsection{PseudoLabeler: Pseudo-labeling raw scans}\label{subsec:PseudoLabeler}

We formulate ground segmentation via the proxy task of estimating a continuous bird's-eye-view (BEV) elevation map
\begin{equation}
    g_\theta: \mathbb{R}^2 \to \mathbb{R},
\end{equation}
parameterized by a multilayer perceptron (MLP). For each point $p_i = (x_i, y_i, z_i)$, the vertical residual $d_i$ is defined as
\begin{equation}
    \Delta d_i = z_i - g_\theta(x_i, y_i).
\end{equation}
If the surface approximation is accurate, points can be segmented using a simple decision rule: points with $\Delta d_i \le D$ are classified as ground, where $D$ is a tolerance threshold accounting for local surface roughness.

To learn $\theta$ directly from raw LiDAR scans, we exploit the geometric prior that the ground is typically the lowest continuous surface in a scene.
Similar to Chodosh et al.~\cite{chodosh2024re}, we fit the elevation map via self-supervised runtime optimization.
We utilize an asymmetric loss to penalize points below the predicted surface quadratically, while applying a Huber penalty~\cite{huber1992robust} to ignore above-ground objects:
\begin{equation}
    \label{eq:chodosh_loss}
    \mathcal{L}(\Delta d_{i}) =
    \begin{cases}
        (\Delta d_{i})^{2}, & \Delta d_{i} < 0,\\
        \mathrm{Huber}(\Delta d_i, \delta), & \Delta d_i \ge 0,
    \end{cases}
\end{equation}
where $\delta$ is the quadratic-to-linear transition threshold, allowing the loss to robustly ignore tall non-ground points.

While conceptually elegant, a naive application of this objective is highly vulnerable to negative sensor noise and often over-fits, leading to severe ground over-segmentation.
To generate high-quality training data, we introduce our \emph{PseudoLabeler}, featuring a three-stage pipeline:

\textbf{Pre-processing.}
Negative noise (\eg multi-path reflections) dominates the asymmetric loss and artificially lowers the elevation map.
To mitigate this, we denoise the raw scan by computing its \qty{0.5}{\percent} lower height quantile and filtering out points falling below this global threshold.

\textbf{Runtime optimization.}
We enhance the MLP architecture with SiLU activations and employ a more resilient optimization strategy.
We optimize Equation~\ref{eq:chodosh_loss} (with $\delta = \qty{0.05}{\meter}$) using AdamW~\cite{loshchilovdecoupled} and a ``ReduceLROnPlateau'' scheduler~\cite{paszke2019pytorch}.
Importantly, to prevent overfitting to non-ground artifacts, we track the exponential moving average (EMA) of the loss and apply early stopping if the EMA loss fails to improve for \num{150} iterations.
Once optimized, we apply a distance threshold $D = \qty{0.40}{\meter}$ to obtain initial labels.

\begin{figure}[t]
  \centering
  \includegraphics[width=\columnwidth]{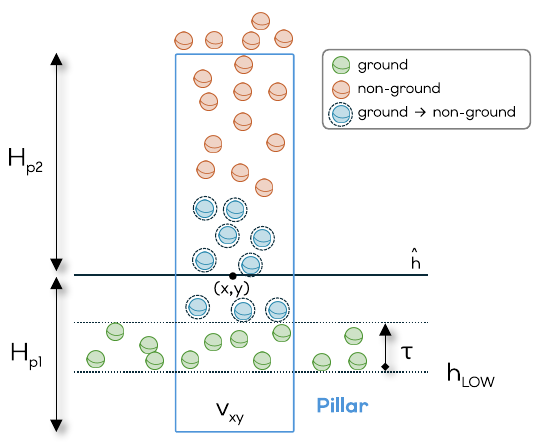}
  \caption{
    \textbf{Post-processing for pseudo-labeling.}
    The point cloud is divided into pillars to recover misclassified non-ground points.
    Points within a vertical window defined by $[ \hat{h} - H_{p1}, \, \hat{h} + H_{p2}]$ around the estimated height $\hat{h} = g (x_i, y_i)$ are analyzed: if a pillar contains \emph{both} ground and non-ground points, all points above the pillar’s lowest point plus margin $\tau$ are reclassified as non-ground.
  }
  \label{fig:PseudoLabeler_post-processing}
\end{figure}

\textbf{Post-processing.}
To address instances where the threshold $D$ over-segments the bottoms of objects (\eg vehicle tires), we apply a pillar-based refinement to recover misclassified non-ground points (illustrated in \Cref{fig:PseudoLabeler_post-processing}). The scan is discretized into \qtyproduct{0.50 x 0.50}{\meter} pillars.
Within a $\pm\qty{1.5}{\meter}$ vertical window around $g_\theta(x, y)$, if a pillar contains both classes, any point exceeding the pillar's minimum elevation by $\tau = \qty{0.05}{\meter}$ is reclassified as non-ground.

\subsection{TerraSeg: Universal ground segmentation}\label{subsec:TerraSeg}

TerraSeg is a real-time, domain-agnostic point cloud model designed for zero-shot LiDAR ground segmentation.
It uses an adapted Point Transformer v3 (PTv3) backbone~\cite{wu2024point} to preserve fine-grained geometric details, followed by a lightweight classification head to predict per-point ground logits.
See part C in \Cref{fig:overview}.
To balance segmentation performance and throughput, we create an accurate Base model (TerraSeg-B) and an efficient Small model (TerraSeg-S).

\textbf{Domain-agnostic.}
To achieve zero-shot generalization, we disable PTv3's dataset-specific normalization, forcing the model to learn universal geometric priors.
To prevent instability from diverse multi-LiDAR batches, we replace 1D Batch Normalization~\cite{ioffe2015batch} with Group Normalization~\cite{wu2018group}.

\textbf{Features.}
To ensure cross-sensor generalization, we restrict the initial point features to a 3-dimensional vector (a constant ones feature, normalized height, and normalized horizontal range), while reserving raw $(x,y,z)$ coordinates strictly for constructing the network's spatial voxel grid.

\textbf{Loss.}
TerraSeg is supervised by the pseudo-labels from PseudoLabeler (see~\ref{subsec:PseudoLabeler}) on OmniLiDAR (see~\ref{subsec:OmniLiDAR}).
We train end-to-end using a per-scan combination of Binary Cross-Entropy (BCE) and a symmetric Lovász-Softmax loss~\cite{berman2018lovasz}:
\begin{equation}
    \mathcal{L} = \mathcal{L}_{\text{BCE}} + \lambda\,\mathcal{L}_{\text{Lov\'asz}},
\end{equation}
where $\lambda$ balances the terms.
We set $\lambda=1.0$.
To adaptively handle class imbalance across diverse scenes, the BCE term applies a dynamic positive-class weight, computed as the ratio of ground to non-ground point fractions tracked via an Exponential Moving Average (EMA) during training.

\section{Experiments}\label{sec:experiments}

\begin{table*}[t]
    \centering
    \caption{
    \textbf{LiDAR ground segmentation on nuScenes~\cite{caesar2020nuscenes} validation split.}
    We train TerraSeg on OmniLiDAR using pseudo-labels from our PseudoLabeler.
    GndNet~\cite{paigwar2020gndnet} is trained on the full nuScenes training split, \ie no sub-sampling.
    The column \emph{Labels} indicates whether manually labeled data is used for training.
    \emph{Prec.} stands for \emph{precision}.
    Best performance in \textbf{bold}, second-best is \underline{underlined}.
    }
    \label{tab:segmentation_nuscenes_val}
    \begin{tabularx}{\textwidth}{>{\raggedright\arraybackslash}p{3.05cm} | >{\centering\arraybackslash}p{0.85cm} | >{\centering\arraybackslash}p{0.95cm} >{\centering\arraybackslash}p{0.95cm} >{\centering\arraybackslash}p{0.95cm} | >{\centering\arraybackslash}p{0.95cm} >{\centering\arraybackslash}p{0.95cm} >{\centering\arraybackslash}p{0.85cm} | >{\centering\arraybackslash}p{1.2cm} | >{\centering\arraybackslash}p{2.4cm}}
        \toprule
        \multicolumn{1}{c}{} & \multicolumn{1}{c}{} & \multicolumn{3}{c}{Ground} & \multicolumn{3}{c}{Non-Ground} & \multicolumn{1}{c}{} & \multicolumn{1}{c}{} \\
        \multicolumn{1}{l}{Method} & \multicolumn{1}{c}{Labels} & \multicolumn{1}{c}{Recall} & \multicolumn{1}{c}{Prec.} & \multicolumn{1}{c}{IoU} & \multicolumn{1}{c}{Recall} & \multicolumn{1}{c}{Prec.} & \multicolumn{1}{c}{IoU} & \multicolumn{1}{c}{mIoU} & \multicolumn{1}{c}{Throughput (\qty{}{\hertz})} \\
        \midrule
        Supervised & \textcolor{citationblue}{\cmark} & 98.15 & 97.73 & 95.96 & 96.97 & 97.53 & 94.65 & 95.31 & \colorbox{SoftGreen}{\phantom{0}28.0} \\
        \midrule
        RANSAC~\cite{fischler1981random} & \textcolor{gray}{\xmark} & 96.02 & 92.56 & 89.14 & 88.19 & 94.60 & 83.97 & 86.55 & \colorbox{SoftGreen}{255.0} \\
        PatchWork++~\cite{lee2022patchwork++} & \textcolor{gray}{\xmark} & 91.46 & 93.73 & 86.19 & 90.23 & 89.29 & 81.42 & 83.80 & \colorbox{SoftGreen}{\phantom{0}30.0} \\
        Chodosh et al.~\cite{chodosh2024re} & \textcolor{gray}{\xmark} & 88.53 & 92.54 & 82.63 & 88.93 & 85.94 & 77.64 & 80.13 & \colorbox{SoftRed}{\phantom{00}0.1} \\
        TRAVEL~\cite{oh2022travel} & \textcolor{gray}{\xmark} & 95.48 & 93.72 & 89.76 & 92.11 & 94.21 & 87.16 & 88.46 & \colorbox{SoftGreen}{\underline{365.7}} \\
        GndNet~\cite{paigwar2020gndnet} & \textcolor{citationblue}{\cmark} & 90.08 & 90.76 & 82.54 & 87.42 & 88.81 & 78.72 & 80.62 & \colorbox{SoftGreen}{\textbf{484.3}} \\
        \midrule
        PseudoLabeler (ours) & \textcolor{gray}{\xmark} & \underline{96.80} & 94.88 & 91.99 & 93.07 & \underline{95.63} & 89.27 & 90.63 & \colorbox{SoftRed}{\phantom{00}0.5} \\        
        TerraSeg-S (ours) & \textcolor{gray}{\xmark} & 95.16 & \textbf{96.96} & \underline{92.40} & \textbf{96.14} & 93.90 & \underline{90.49} & \underline{91.45} & \colorbox{SoftGreen}{\phantom{0}49.8} \\
        TerraSeg-B (ours) & \textcolor{gray}{\xmark} & \textbf{97.88} & \underline{95.43} & \textbf{93.50} & \underline{93.95} & \textbf{97.17} & \textbf{91.45} & \textbf{92.47} & \colorbox{SoftGreen}{\phantom{0}28.0} \\
        \bottomrule
    \end{tabularx}
\end{table*}

We conduct a comprehensive evaluation of our proposed framework, focusing on the task of ground segmentation.
Section~\ref{subsection_experimental_setup} describes the experimental setup.
Subsequently, we present quantitative comparisons against state-of-the-art baselines in Section~\ref{subsection_ground_segmentation}.
Finally, Section~\ref{subsec:ablation} provides ablation studies analyzing our pseudo-labeling components (\ie pre-processing, post-processing), hyperparameter sensitivity, and model performance across varying terrain flatness.

\subsection{Experimental setup}\label{subsection_experimental_setup}

\textbf{Dataset.}
We train TerraSeg on OmniLiDAR, our unified dataset of nearly \num{22} million raw scans from \num{12} public autonomous driving benchmarks, encompassing 15 distinct LiDAR hardware models.
PseudoLabeler provides all training labels without human annotation.
We evaluate on the held-out validation splits of nuScenes, SemanticKITTI, and Waymo Perception, treating them as our labeled test sets.

\textbf{Metrics.}
We evaluate ground segmentation using per-class \emph{recall}, \emph{precision}, and \emph{intersection-over-union (IoU)} for the ground and non-ground classes, as well as their \emph{mean IoU (mIoU)}.
These capture both omission and commission errors.
All evaluation is per-frame; no temporal fusion.

\textbf{Baselines.}
We benchmark against (1) classical geometry methods (RANSAC~\cite{fischler1981random}, PatchWork++~\cite{lee2022patchwork++}, TRAVEL~\cite{oh2022travel}), (2) self-supervised runtime optimization (Chodosh et al.~\cite{chodosh2024re}), and a supervised model (GndNet~\cite{paigwar2020gndnet}).
To establish an upper performance bound, we also evaluate a supervised oracle baseline: TerraSeg-B trained on labeled data from nuScenes~\cite{caesar2020nuscenes}, SemanticKITTI~\cite{behley2019semantickitti}, Waymo Perception~\cite{sun2020scalability}, AevaScenes~\cite{narasimhan2025aevascenes}, and PandaSet~\cite{xiao2021pandaset}.

\textbf{Implementation details.}
TerraSeg is implemented in PyTorch~\cite{paszke2019pytorch} using the PTv3~\cite{wu2024point} backbone.
Inputs are 3D coordinates voxelized at \qty{0.05}{\meter}, augmented with 3-dimensional features (see Section~\ref{subsec:TerraSeg}).
During training, we apply SE(3) augmentations, jitter, and point dropout.
We optimize a per-scan combination of class-balanced Binary Cross-Entropy (BCE) and symmetric Lovász-Softmax loss.
We train using AdamW~\cite{loshchilovdecoupled} (learning rate $2\times10^{-3}$, weight decay $5\times10^{-3}$, gradient clipping at \num{1.0}) with a one-epoch linear warm-up followed by cosine decay.
We use \num{4} scans per dataloader step and apply gradient accumulation to achieve an effective batch size of \num{256}.
To mitigate dataset imbalance, we utilize a custom data sampler that balances LiDAR sweeps across datasets according to defined probabilities, setting a custom epoch length of \num{20,000} LiDAR samples.
We release our \href{https://www.github.com/TedLentsch/TerraSeg}{code} under the \emph{Apache 2.0 license}.

\subsection{LiDAR ground segmentation}\label{subsection_ground_segmentation}

\begin{table*}[t]
    \centering
    \caption{
    \textbf{LiDAR ground segmentation on SemanticKITTI~\cite{behley2019semantickitti} validation split.}
    We train TerraSeg on OmniLiDAR using pseudo-labels from our PseudoLabeler.
    GndNet~\cite{paigwar2020gndnet} is trained on the full SemanticKITTI training split, \ie no sub-sampling.
    The column \emph{Labels} indicates whether manually labeled data is used for training.
    \emph{Prec.} stands for \emph{precision}.
    Best performance in \textbf{bold}, second-best is \underline{underlined}.
    }
    \label{tab:segmentation_kitti_val}
    \begin{tabularx}{\textwidth}{>{\raggedright\arraybackslash}p{3.05cm} | >{\centering\arraybackslash}p{0.85cm} | >{\centering\arraybackslash}p{0.95cm} >{\centering\arraybackslash}p{0.95cm} >{\centering\arraybackslash}p{0.95cm} | >{\centering\arraybackslash}p{0.95cm} >{\centering\arraybackslash}p{0.95cm} >{\centering\arraybackslash}p{0.85cm} | >{\centering\arraybackslash}p{1.2cm} | >{\centering\arraybackslash}p{2.4cm}}
        \toprule
        \multicolumn{1}{c}{} & \multicolumn{1}{c}{} & \multicolumn{3}{c}{Ground} & \multicolumn{3}{c}{Non-Ground} & \multicolumn{1}{c}{} & \multicolumn{1}{c}{} \\
        \multicolumn{1}{l}{Method} & \multicolumn{1}{c}{Labels} & \multicolumn{1}{c}{Recall} & \multicolumn{1}{c}{Prec.} & \multicolumn{1}{c}{IoU} & \multicolumn{1}{c}{Recall} & \multicolumn{1}{c}{Prec.} & \multicolumn{1}{c}{IoU} & \multicolumn{1}{c}{mIoU} & \multicolumn{1}{c}{Throughput (\qty{}{\hertz})} \\
        \midrule
        Supervised & \textcolor{citationblue}{\cmark} & 96.70 & 97.26 & 94.13 & 97.69 & 97.21 & 95.03 & 94.58 & \colorbox{SoftOrange}{\phantom{00}9.7} \\        
        \midrule
        RANSAC~\cite{fischler1981random} & \textcolor{gray}{\xmark} & 97.14 & 79.61 & 77.79 & 78.94 & 97.12 & 77.13 & 77.46 & \colorbox{SoftGreen}{\underline{255.0}} \\
        PatchWork++~\cite{lee2022patchwork++} & \textcolor{gray}{\xmark} & 96.57 & 91.72 & 88.82 & 92.15 & 97.04 & 89.64 & 89.23 & \colorbox{SoftGreen}{\phantom{0}30.0} \\
        Chodosh et al.~\cite{chodosh2024re} & \textcolor{gray}{\xmark} & 95.52 & 83.53 & 80.38 & 83.88 & 95.82 & 80.91 & 80.64 & \colorbox{SoftRed}{\phantom{00}0.1} \\
        TRAVEL~\cite{oh2022travel} & \textcolor{gray}{\xmark} & 95.38 & 89.77 & 86.28 & 92.72 & 96.98 & 90.09 & 88.19 & \colorbox{SoftGreen}{\phantom{0}85.1} \\
        GndNet~\cite{paigwar2020gndnet} & \textcolor{citationblue}{\cmark} & \textbf{99.72} & 80.72 & 80.54 & 83.41 & \textbf{99.70} & 83.19 & 81.87 & \colorbox{SoftGreen}{\textbf{394.0}} \\
        \midrule
        PseudoLabeler (ours) & \textcolor{gray}{\xmark} & 96.53 & 94.52 & 91.41 & 95.25 & 97.00 & 92.53 & 91.97 & \colorbox{SoftRed}{\phantom{00}0.3} \\        
        TerraSeg-S (ours) & \textcolor{gray}{\xmark} & 95.75 & \textbf{96.33} & \underline{92.38} & \textbf{97.01} & 96.53 & \underline{93.73} & \underline{93.06} & \colorbox{SoftGreen}{\phantom{0}17.4} \\
        TerraSeg-B (ours) & \textcolor{gray}{\xmark} & \underline{97.65} & \underline{95.10} & \textbf{92.97} & \underline{95.87} & \underline{98.03} & \textbf{94.06} & \textbf{93.52} & \colorbox{SoftOrange}{\phantom{00}9.7} \\
        \bottomrule
    \end{tabularx}
\end{table*}

\begin{table*}[t]
    \centering
    \caption{
    \textbf{LiDAR ground segmentation on Waymo Perception~\cite{sun2020scalability} validation split.}
    We train TerraSeg on OmniLiDAR using pseudo-labels from our PseudoLabeler.
    GndNet~\cite{paigwar2020gndnet} is trained on the full Waymo training split, \ie no sub-sampling.
    The column \emph{Labels} indicates whether manually labeled data is used for training.
    \emph{Prec.} stands for \emph{precision}.
    Best performance in \textbf{bold}, second-best is \underline{underlined}.
    }
    \label{tab:segmentation_wop_val}
    \begin{tabularx}{\textwidth}{>{\raggedright\arraybackslash}p{3.05cm} | >{\centering\arraybackslash}p{0.85cm} | >{\centering\arraybackslash}p{0.95cm} >{\centering\arraybackslash}p{0.95cm} >{\centering\arraybackslash}p{0.95cm} | >{\centering\arraybackslash}p{0.95cm} >{\centering\arraybackslash}p{0.95cm} >{\centering\arraybackslash}p{0.85cm} | >{\centering\arraybackslash}p{1.2cm} | >{\centering\arraybackslash}p{2.4cm}}
        \toprule
        \multicolumn{1}{c}{} & \multicolumn{1}{c}{} & \multicolumn{3}{c}{Ground} & \multicolumn{3}{c}{Non-Ground} & \multicolumn{1}{c}{} & \multicolumn{1}{c}{} \\
        \multicolumn{1}{l}{Method} & \multicolumn{1}{c}{Labels} & \multicolumn{1}{c}{Recall} & \multicolumn{1}{c}{Prec.} & \multicolumn{1}{c}{IoU} & \multicolumn{1}{c}{Recall} & \multicolumn{1}{c}{Prec.} & \multicolumn{1}{c}{IoU} & \multicolumn{1}{c}{mIoU} & \multicolumn{1}{c}{Throughput (\qty{}{\hertz})} \\
        \midrule
        Supervised & \textcolor{citationblue}{\cmark} & 98.05 & 98.22 & 96.34 & 98.81 & 98.69 & 97.53 & 96.94 & \colorbox{SoftOrange}{\phantom{0}10.1} \\
        \midrule
        RANSAC~\cite{fischler1981random} & \textcolor{gray}{\xmark} & 91.86 & 76.14 & 71.32 & 81.64 & 94.02 & 77.61 & 74.47 & \colorbox{SoftGreen}{\underline{255.0}} \\
        PatchWork++~\cite{lee2022patchwork++} & \textcolor{gray}{\xmark} & 69.16 & 92.00 & 65.23 & 96.16 & 83.02 & 80.36 & 72.80 & \colorbox{SoftGreen}{\phantom{0}30.0} \\
        Chodosh et al.~\cite{chodosh2024re} & \textcolor{gray}{\xmark} & \underline{97.09} & 84.84 & 82.74 & 88.94 & \underline{97.95} & 87.31 & 85.03 & \colorbox{SoftRed}{\phantom{00}0.1} \\
        TRAVEL~\cite{oh2022travel} & \textcolor{gray}{\xmark} & 96.20 & 91.63 & 88.55 & 94.81 & 96.98 & 92.24 & 90.40 & \colorbox{SoftGreen}{\phantom{0}76.9} \\
        GndNet~\cite{paigwar2020gndnet} & \textcolor{citationblue}{\cmark} & 83.57 & 65.79 & 57.64 & 74.31 & 85.65 & 66.01 & 61.83 & \colorbox{SoftGreen}{\textbf{356.0}} \\
        \midrule
        PseudoLabeler (ours) & \textcolor{gray}{\xmark} & 95.57 & \underline{95.65} & 91.60 & \underline{97.09} & 97.04 & 94.30 & 92.95 & \colorbox{SoftRed}{\phantom{00}0.3} \\
        TerraSeg-S (ours) & \textcolor{gray}{\xmark} & 96.33 & \textbf{96.25} & \underline{92.84} & \textbf{97.60} & 97.66 & \underline{95.37} & \underline{94.11} & \colorbox{SoftGreen}{\phantom{0}17.4} \\
        TerraSeg-B (ours) & \textcolor{gray}{\xmark} & \textbf{97.97} & 95.10 & \textbf{93.26} & 96.78 & \textbf{98.68} & \textbf{95.54} & \textbf{94.40} & \colorbox{SoftOrange}{\phantom{0}10.1} \\
        \bottomrule
    \end{tabularx}
    \vspace{-1mm}
\end{table*}

Tables~\ref{tab:segmentation_nuscenes_val}, \ref{tab:segmentation_kitti_val}, and \ref{tab:segmentation_wop_val} summarize the results of ground segmentation across the three evaluation benchmarks~\cite{caesar2020nuscenes, behley2019semantickitti, sun2020scalability}.
We observe that while geometric methods struggle with non-planar terrain and supervised methods overfit to specific sensor configurations, our domain-agnostic, self-supervised model TerraSeg robustly bridges this generalization gap.

\textbf{Handcrafted baselines.}
Classical geometric methods like RANSAC~\cite{fischler1981random} and PatchWork++~\cite{lee2022patchwork++} remain competitive in structured urban environments but exhibit clear limitations.
RANSAC systematically trims the lower parts of objects, while PatchWork++ excels on locally planar scenes but suffers from under-segmentation in dense sensing, \ie low recall on Waymo Perception.
TRAVEL~\cite{oh2022travel} emerges as the strongest handcrafted baseline, achieving \num{88.46} mIoU on nuScenes and \num{90.40} mIoU on Waymo Perception, yet it lacks the adaptability of learning-based models.

\textbf{Self-supervised and Supervised baselines.}
Chodosh et al.~\cite{chodosh2024re} achieve strong performance on Waymo Perception (\num{85.03} mIoU), the most non-flat dataset, because its piece-wise linear surface naturally accommodates terrain elevation changes better than global planar models.
However, it suffers from the noisy LiDAR returns on nuScenes and SemanticKITTI.
In contrast, the supervised GndNet~\cite{paigwar2020gndnet} fails to generalize across LiDAR configurations; constrained by its handcrafted pre-processing, its performance severely degrades on Waymo Perception to \num{61.83} mIoU.

\textbf{TerraSeg.}
Across all benchmarks, TerraSeg-B (46M parameters) sets a new state-of-the-art despite using no manual labels, achieving \num{92.47} mIoU on nuScenes,\num{93.52} mIoU on SemanticKITTI, and \num{94.40} mIoU on Waymo.
It outperforms all baselines on nearly all metrics, with the sole exception of GndNet on SemanticKITTI, which attains a higher ground recall (\num{99.72}) and non-ground precision (\num{99.70}) but TerraSeg has significantly higher mIoU.

Tables~\ref{tab:segmentation_nuscenes_val}--\ref{tab:segmentation_wop_val} also indicate throughputs: \colorbox{SoftRed}{\qty{<5}{\hertz}} (offline), \colorbox{SoftOrange}{\qtyrange{5}{15}{\hertz}} (near real-time), and \colorbox{SoftGreen}{\qty{>15}{\hertz}} (real-time). 
Crucially, disentangling the label generation from inference allows us to balance segmentation performance and speed.
While PseudoLabeler generates high-quality labels offline (\qtyrange{0.3}{0.5}{\hertz}), our deployable models achieve practical speeds on an NVIDIA A100 GPU. 
TerraSeg-B runs near real-time (\qtyrange{10}{28}{\hertz}), and the efficient TerraSeg-S (12M parameters) maintains real-time throughputs (\qtyrange{17}{50}{\hertz}) as the second-best overall model based on mIoU.

\begin{table}[t]
    \centering
    \caption{
    \textbf{PseudoLabeler ablation study I.}
    We progressively add pre-processing and post-processing to the base model.
    We report mIoU on nuScenes (NS), SemanticKITTI (SK), and Waymo Perception (WP) validation splits.
    The column \emph{TP} indicates \emph{Throughput (Hz)}.
    Best performance in \textbf{bold}, second-best is \underline{underlined}.
    }
    \label{tab:ablation_pseudolabeler1}
    \begin{tabularx}{\linewidth}{
        l
        | >{\centering\arraybackslash}X
          >{\centering\arraybackslash}X
          >{\centering\arraybackslash}X
        | >{\centering\arraybackslash}X
    }
        \toprule
        Method & NS & SK & WP & TP \\
        \midrule
        Chodosh et al.~\cite{chodosh2024re} & 80.13 & 80.64 & 85.03 & 0.1 \\
        \midrule
        PseudoLabeler Base & 79.40 & 80.01 & 85.00 & \textbf{0.3} \\
        + Pre-processing & \underline{86.38} & \underline{83.82} & \underline{85.47} & \textbf{0.3} \\
        + Post-processing & \textbf{90.63} & \textbf{91.97} & \textbf{92.95} & \textbf{0.3} \\
        \bottomrule
    \end{tabularx}
\end{table}

\subsection{Ablation studies}\label{subsec:ablation}

In \Cref{tab:ablation_pseudolabeler1}, we ablate PseudoLabeler's key design choices.
Starting from the base, \ie runtime optimization without filtering (see Section~\ref{subsec:PseudoLabeler}), adding region-wise pre-processing substantially improves mIoU (+\num{6.98} on nuScenes) by removing undesired noise artifacts.
Furthermore, pillar-based post-processing (see \Cref{fig:PseudoLabeler_post-processing}) to recover misclassified non-ground points provides a consistent, significant boost across all datasets (ranging from +\num{4.25} to +\num{8.15} mIoU).

\Cref{tab:ablation_pseudolabeler2} analyzes the sensitivity of this post-processing step from PseudoLabeler.
The optimal configuration is $v_{xy} =$ \qty{0.50}{\meter} and $\tau =$ \qty{0.05}{\meter}, yielding the highest average mIoU of \num{91.85}.
In general, increasing the recovery threshold $\tau$ degrades performance, while pillar size $v_{xy}$ requires a balanced mid-range value to function effectively.

\Cref{tab:ablation_terraseg1} evaluates TerraSeg-B across flat and non-flat terrain based on the ground height standard deviation ($\sigma$).
TerraSeg-B demonstrates strong local adaptability, maintaining more than \num{87} mIoU even in highly non-planar scenes (\ie nuScenes non-flat).
Furthermore, dataset roughness directly exposes the limits of global planar priors.
As median $\sigma$ increases from nuScenes (15.19 cm) to SemanticKITTI (16.10 cm) and Waymo Perception (18.38 cm), TerraSeg's mIoU advantage over RANSAC~\cite{fischler1981random} scales proportionally (+5.92, +16.06, and +19.93, respectively).

\begin{table}[t]
    \centering
    \caption{
    \textbf{PseudoLabeler ablation study II.}
    We ablate the pillar size $v_{xy}$ and recovery threshold $\tau$ for post-processing.
    We report mIoU on nuScenes (NS), SemanticKITTI (SK), and Waymo Perception (WP) validation splits.
    The column \emph{Avg} indicates \emph{Average mIoU}.
    Best performance in \textbf{bold}, second-best is \underline{underlined}.
    }
    \label{tab:ablation_pseudolabeler2}
    \begin{tabularx}{\linewidth}{
          >{\centering\arraybackslash}X
          >{\centering\arraybackslash}X
        | >{\centering\arraybackslash}X
          >{\centering\arraybackslash}X
          >{\centering\arraybackslash}X
        | >{\centering\arraybackslash}X
    }
        \toprule
        $v_{xy}$~(m) & $\tau$ (m)  & NS & SK & WP & Avg\\
        \midrule
        0.25 & 0.05 & 89.96 & 91.11 & 92.17 & 91.08 \\
        0.25 & 0.10 & 89.79 & 89.87 & 91.66 & 90.44 \\
        0.25 & 0.20 & 88.98 & 86.98 & 89.82 & 88.59 \\
        \midrule
        0.50 & 0.05 & \textbf{90.63} & \textbf{91.97} & \textbf{92.95} & \textbf{91.85} \\
        0.50 & 0.10 & \underline{90.50} & 91.60 & 92.69 & 91.59 \\
        0.50 & 0.20 & 89.75 & 88.76 & 90.76 & 89.76 \\
        \midrule
        1.00 & 0.05 & 89.94 & 89.68 & 92.27 & 90.63 \\
        1.00 & 0.10 & \underline{90.50} & \underline{91.92} & \underline{92.90} & \underline{91.77} \\
        1.00 & 0.20 & 90.24 & 90.42 & 91.65 & 90.77 \\
        \bottomrule
    \end{tabularx}
\end{table}

\begin{table}[t]
    \centering
    \caption{
    \textbf{TerraSeg ablation study I.}
    We report mIoU across datasets by partitioning validation scans into flat and non-flat, based on the per-scan standard deviation of ground elevation.
    }
    \label{tab:ablation_terraseg1}
    \begin{tabularx}{\linewidth}{
        l
        | >{\centering\arraybackslash}X
          >{\centering\arraybackslash}X
    }
        \toprule
        Dataset & $\sigma \leq$ \qty{0.40}{\meter} & $\sigma >$ \qty{0.40}{\meter} \\
        \midrule
        nuScenes~\cite{caesar2020nuscenes} & 93.12 & 87.81 \\
        SemanticKITTI~\cite{behley2019semantickitti} & 93.37 & 92.60 \\
        Waymo Perception~\cite{sun2020scalability} & 95.10 & 90.82 \\
        \bottomrule
    \end{tabularx}
    \vspace{-2mm}
\end{table}

\section{Conclusion}\label{sec:conclusion}

We presented a novel, self-supervised method for domain-agnostic LiDAR ground segmentation.
Our approach is built on three core components: OmniLiDAR, a diverse, large-scale aggregated dataset; PseudoLabeler, which generates reliable pseudo-labels via self-supervised per-scan optimization; and TerraSeg, a real-time model trained on this corpus.
Extensive experiments on the nuScenes~\cite{caesar2020nuscenes}, SemanticKITTI~\cite{behley2019semantickitti}, and Waymo Perception~\cite{sun2020scalability} datasets demonstrate that TerraSeg achieves state-of-the-art performance without manual annotations.
Furthermore, ablation studies validate the integration of domain knowledge within PseudoLabeler's pre- and post-processing stages.

\textbf{Future work.}
Promising directions include: (1) leveraging temporal context to enhance pseudo-labeling in sparse regions; (2) iteratively self-training TerraSeg to exploit its pseudo-label denoising capabilities; and (3) initializing the backbone with foundation models (\eg Sonata~\cite{wu2025sonata}).

\section*{Acknowledgments}
This research has been conducted as part of the EVENTS project, which is funded by the European Union, under grant agreement No 101069614.
Views and opinions expressed are, however, those of the author(s) only and do not necessarily reflect those of the European Union or European Commission.
Neither the European Union nor the granting authority can be held responsible for them.
This work has also been supported by project PID2024-161576OB-I00, funded by Spanish MICIU/AEI/10.13039/501100011033 and co-funded by the European Regional Development Fund (ERDF, ``A way of making Europe'').

{
    \small
    \bibliographystyle{ieeenat_fullname}
    \bibliography{main}
}

\clearpage
\appendix
\section{OmniLiDAR details}

\subsection{Dataset sampling distribution}

As mentioned in the main manuscript, the datasets within OmniLiDAR vary drastically in size.
For instance, Argoverse 2 Lidar~\cite{wilson2023argoverse} contains nearly 12 million raw scans, whereas smaller datasets like View-of-Delft~\cite{palffy2022multi} or SemanticKITTI~\cite{behley2019semantickitti} contain significantly fewer.
To prevent the model from biasing toward the largest data collections and to ensure balanced cross-sensor generalization, we employ a custom weighted sampling strategy during training.

We group the datasets into specific sampling buckets and assign a fixed target probability to each.
The distribution is designed to prioritize our primary evaluation domains while capping the influence of massive collections:

\begin{itemize}
    \item \textbf{Primary datasets (60\% total):} The datasets corresponding to our main evaluation benchmarks are heavily prioritized. We allocate \qty{20}{\percent} to \texttt{nuScenes\_main}, \qty{20}{\percent} to \texttt{WaymoPerception\_main}, and \qty{20}{\percent} to \texttt{KITTI\_main}. 
    \begin{itemize}
        \item For \texttt{KITTI\_main}, we merge SemanticKITTI~\cite{behley2019semantickitti} and KITTI-360~\cite{liao2022kitti} into a single group, as they share highly similar environments and the exact same hardware configuration (Velodyne HDL-64E).
        \item For \texttt{WaymoPerception\_main}, we strictly route scans from the primary roof-mounted panoramic sensor (\texttt{TOP}) to this group~\cite{sun2020scalability}. The shorter-range secondary sensors are routed to a generic fallback group.
    \end{itemize}
    \item \textbf{Massive datasets (15\% total):} To prevent domination during training, we cap the largest datasets, \ie Argoverse 2 Lidar, ONCE, and Zenseact Open Dataset (ZOD), at \qty{5}{\percent} each.
    \item \textbf{Other datasets (22.5\% total):} We assign \qty{5}{\percent} each to AevaScenes~\cite{narasimhan2025aevascenes}, Lyft~\cite{kesten2019lyft}, PandaSet~\cite{xiao2021pandaset}, and TruckScenes~\cite{fent2024man}. Because View-of-Delft~\cite{palffy2022multi} is a relatively small dataset, we assign it a lower sampling weight of \qty{2.5}{\percent}.
    \item \textbf{Fallback group (2.5\% total):} An \texttt{else} group is allocated the remaining \qty{2.5}{\percent}. This group catches Waymo's non-TOP secondary sensors~\cite{sun2020scalability} and any other unspecified scan data.
\end{itemize}

By utilizing these predefined probabilities, our dataloader ensures that each epoch (defined as \num{20000} LiDAR samples) exposes the network to a consistently diverse, domain-balanced batch of geometries.

\subsection{Overview of unique LiDAR sensors}

In Table \ref{tab:unique_lidars_appendix}, we detail the 15 distinct LiDAR hardware models aggregated within the OmniLiDAR dataset.
This hardware diversity is fundamental to achieving TerraSeg's domain-agnostic generalization.
The table illustrates a broad spectrum of sensor modalities, ranging from classic high-resolution spinning LiDARs (\eg Velodyne HDL-64E, Hesai Pandar64) to specialized configurations such as roof-mounted panoramic sensors, near-field short-range LiDARs, and Frequency-Modulated Continuous Wave (FMCW) sensors (Aeva).
By exposing our self-supervised model to this broad variation in scan patterns, point densities, and vertical fields of view during training, we encourage the network to learn universal geometric priors rather than overfitting to sensor-specific artifacts.

\subsection{OmniLiDAR edge cases}

During the curation of the OmniLiDAR dataset, we standardize and process raw scans to ensure consistency across the \num{12} datasets.
To maintain a robust data loading pipeline, we addressed the following dataset-specific edge cases:

\noindent\textbf{1. Missing returns in Argoverse 2 Lidar:}
Occasionally, the secondary LiDAR sensors in this dataset yield an empty point cloud.
Because our data loading protocol expects continuous files for the curated \qty{0.2}{\hertz} sequences, we generate placeholder empty files for these instances.
During training, if the dataloader encounters an empty file, it dynamically resamples to retrieve a valid point cloud.
Out of the \num{240000} processed files for this dataset, \num{236904} contain valid sensor returns, which is accurately reflected in the aggregate statistics.

\noindent\textbf{2. Frame selection in Zenseact Open Dataset:}
Zenseact provides curated frames structured as two-second snippets containing past, key, and future scans.
To align this with our single-scan, per-frame processing strategy, we exclusively extract the central key scan from each frame.
Because we subsequently downsample the overall sequences to \qty{0.2}{\hertz} (one scan every 5 seconds), we naturally avoid any temporal overlap or redundancy between the 2-second snippets.

\section{Qualitative results}

\subsection{Success cases}

\begin{figure*}[h]
    \centering
    \includegraphics[width=0.32\linewidth]{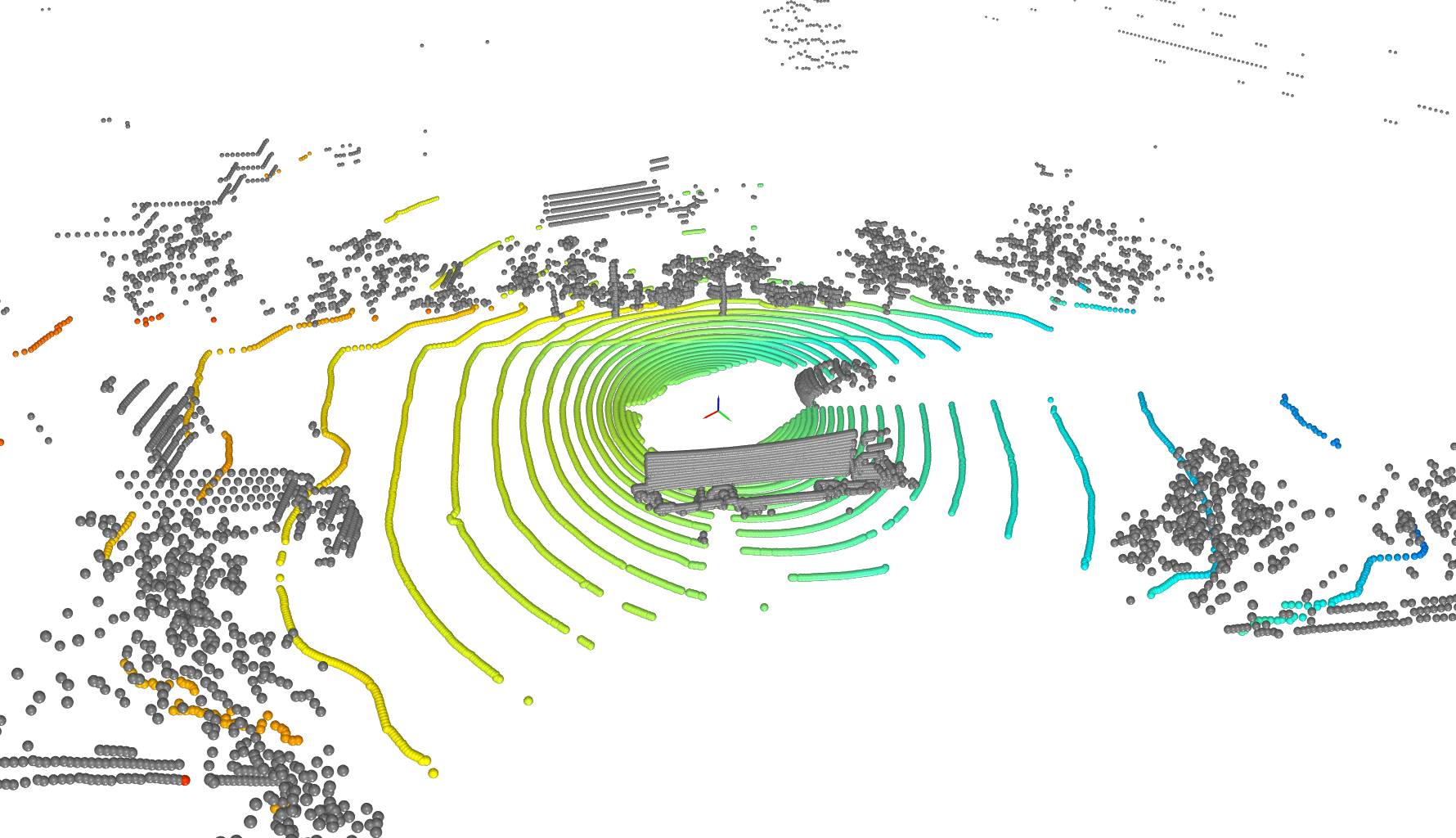}
    \includegraphics[width=0.32\linewidth]{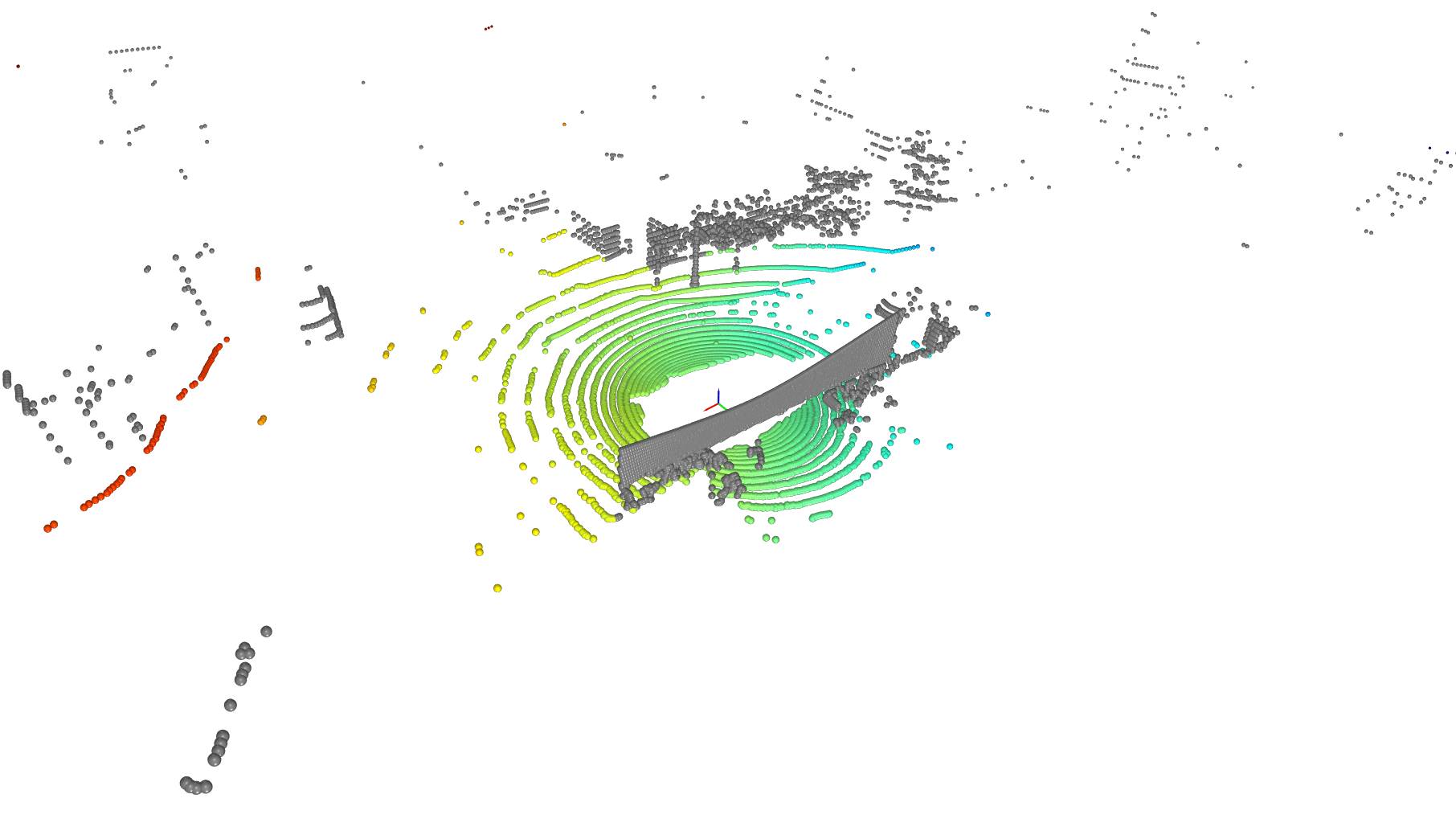}
    \includegraphics[width=0.32\linewidth]{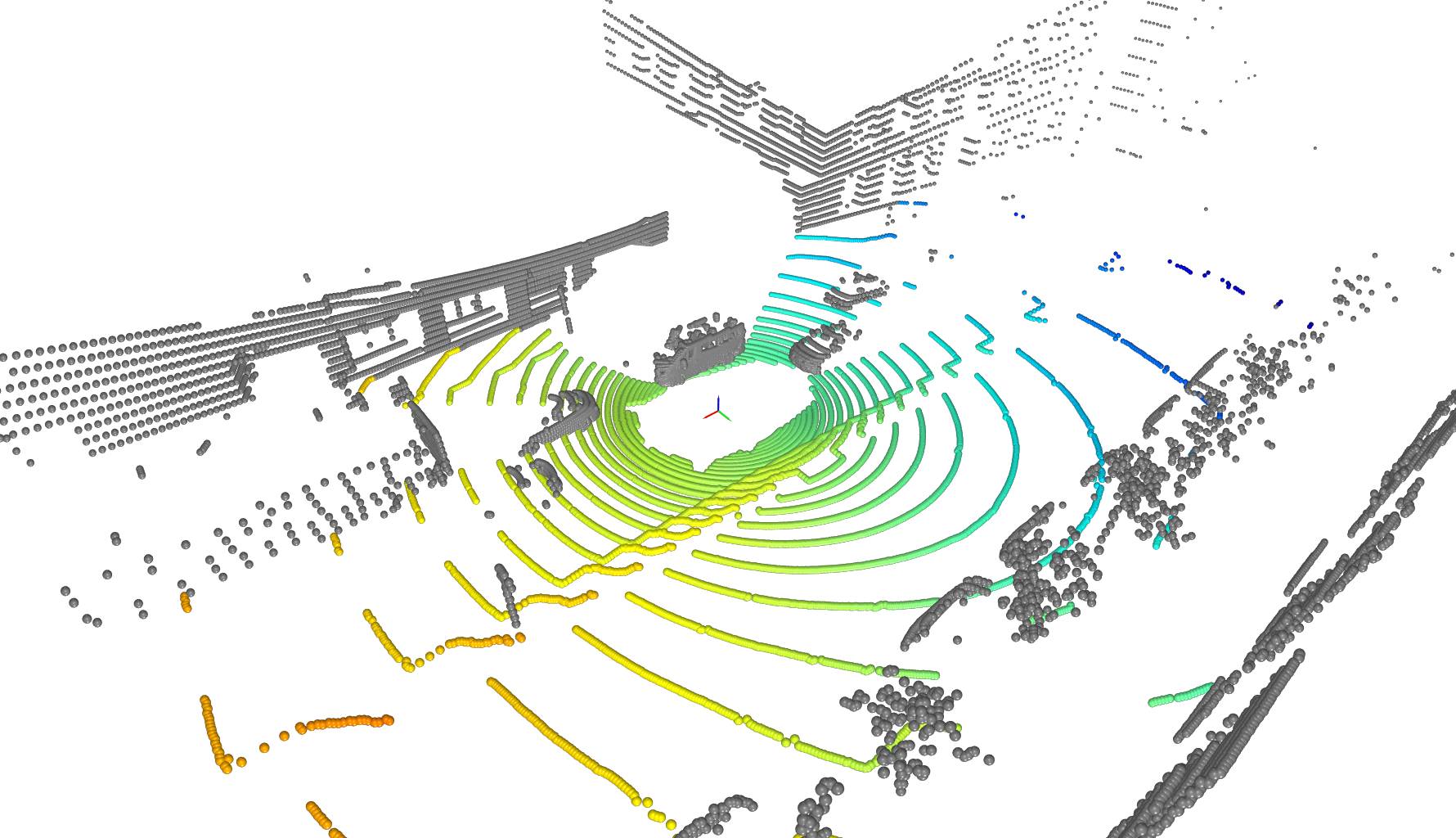}\\
    \vspace{2mm}
    \includegraphics[width=0.32\linewidth]{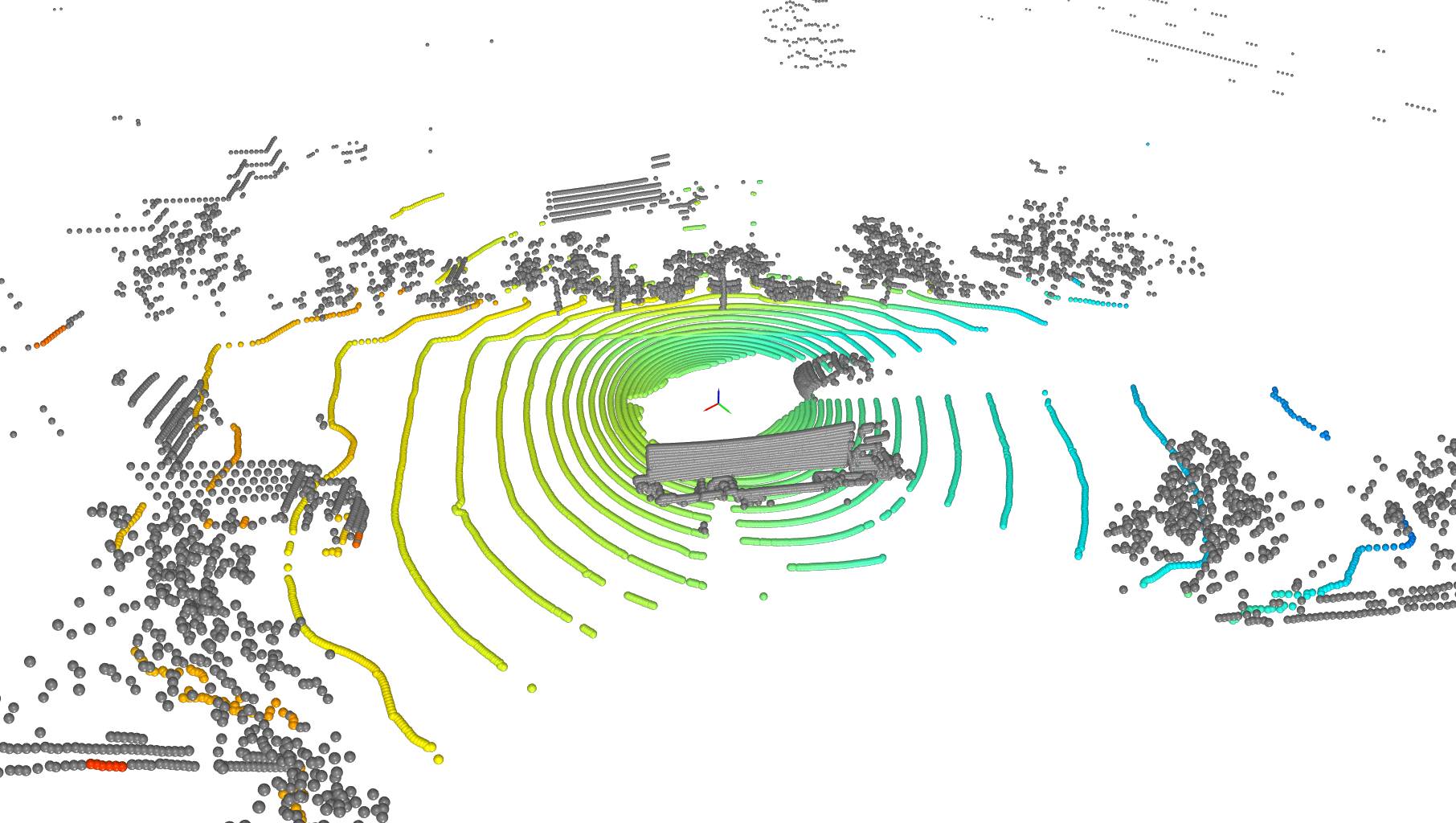}
    \includegraphics[width=0.32\linewidth]{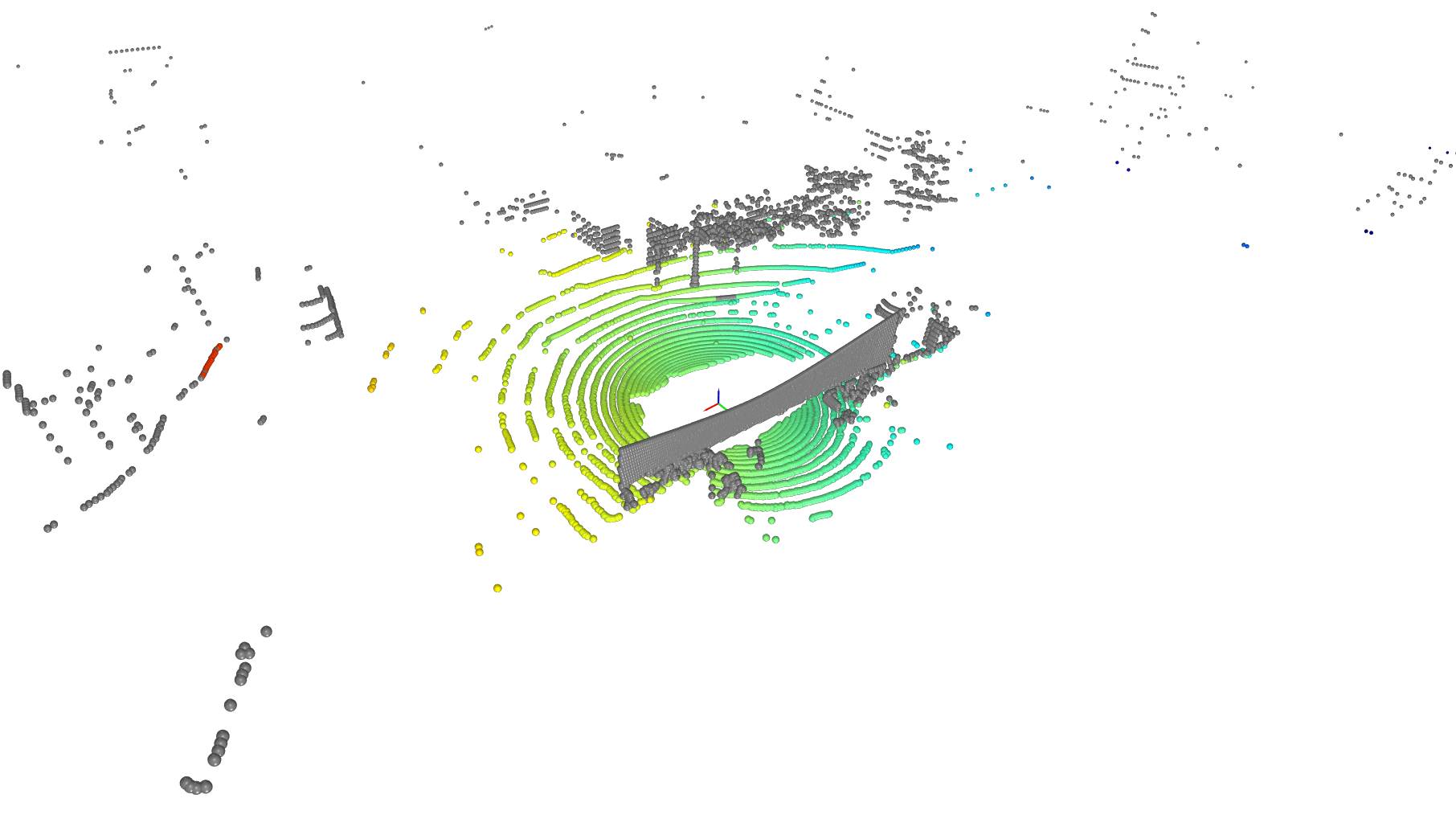}
    \includegraphics[width=0.32\linewidth]{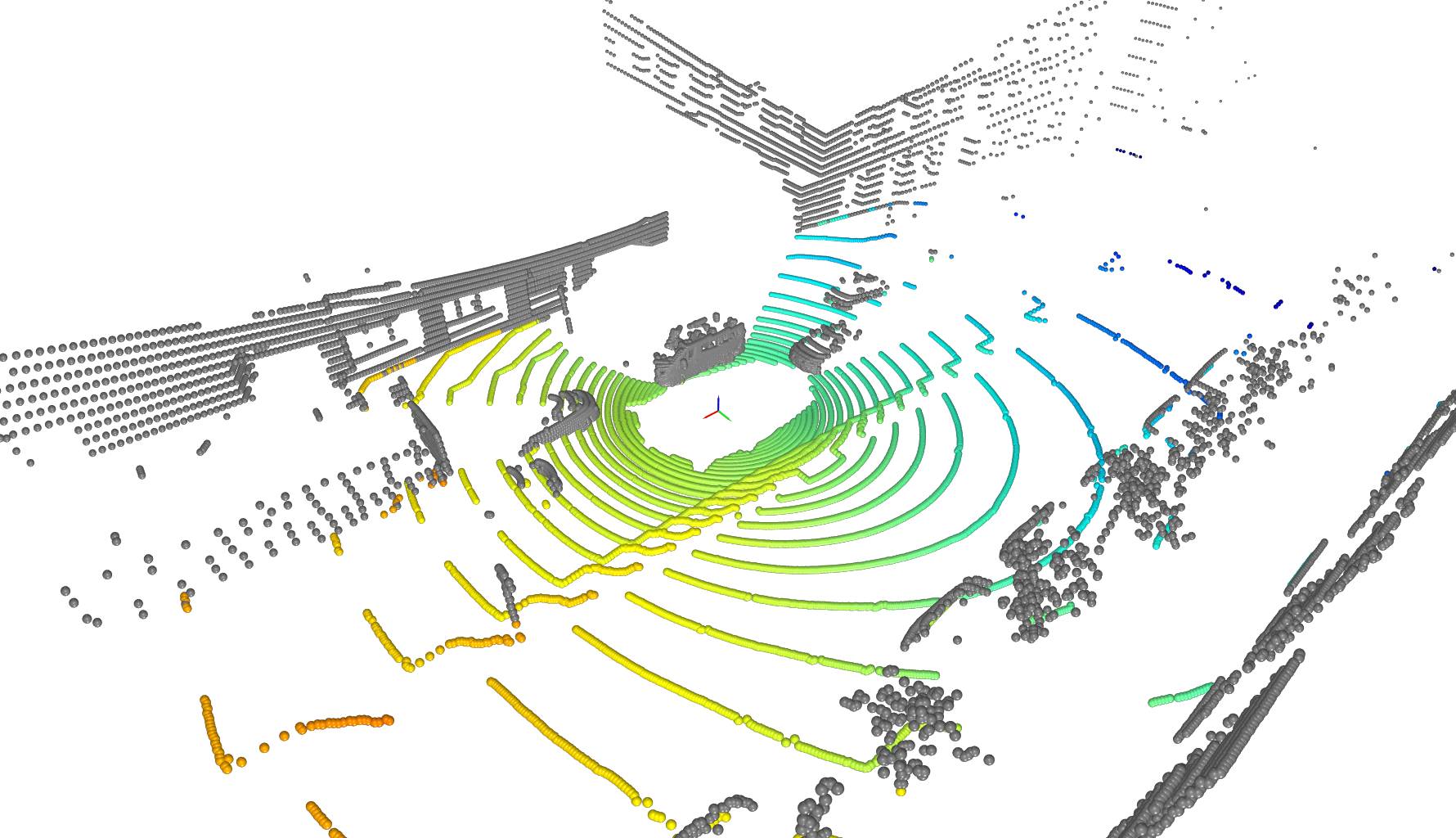}
    \caption{
    \textbf{Qualitative success cases on nuScenes dataset~\cite{caesar2020nuscenes}.} The top row shows ground truth labels, while the bottom row displays TerraSeg predictions. Ground points are color-coded by elevation; non-ground points are rendered in gray. From left to right, columns correspond to scan IDs \num{1787}, \num{3238}, and \num{3479}.}
    \label{fig:supp_success_nuscenes}
\end{figure*}

\begin{figure*}[h]
    \centering
    \includegraphics[width=0.32\linewidth]{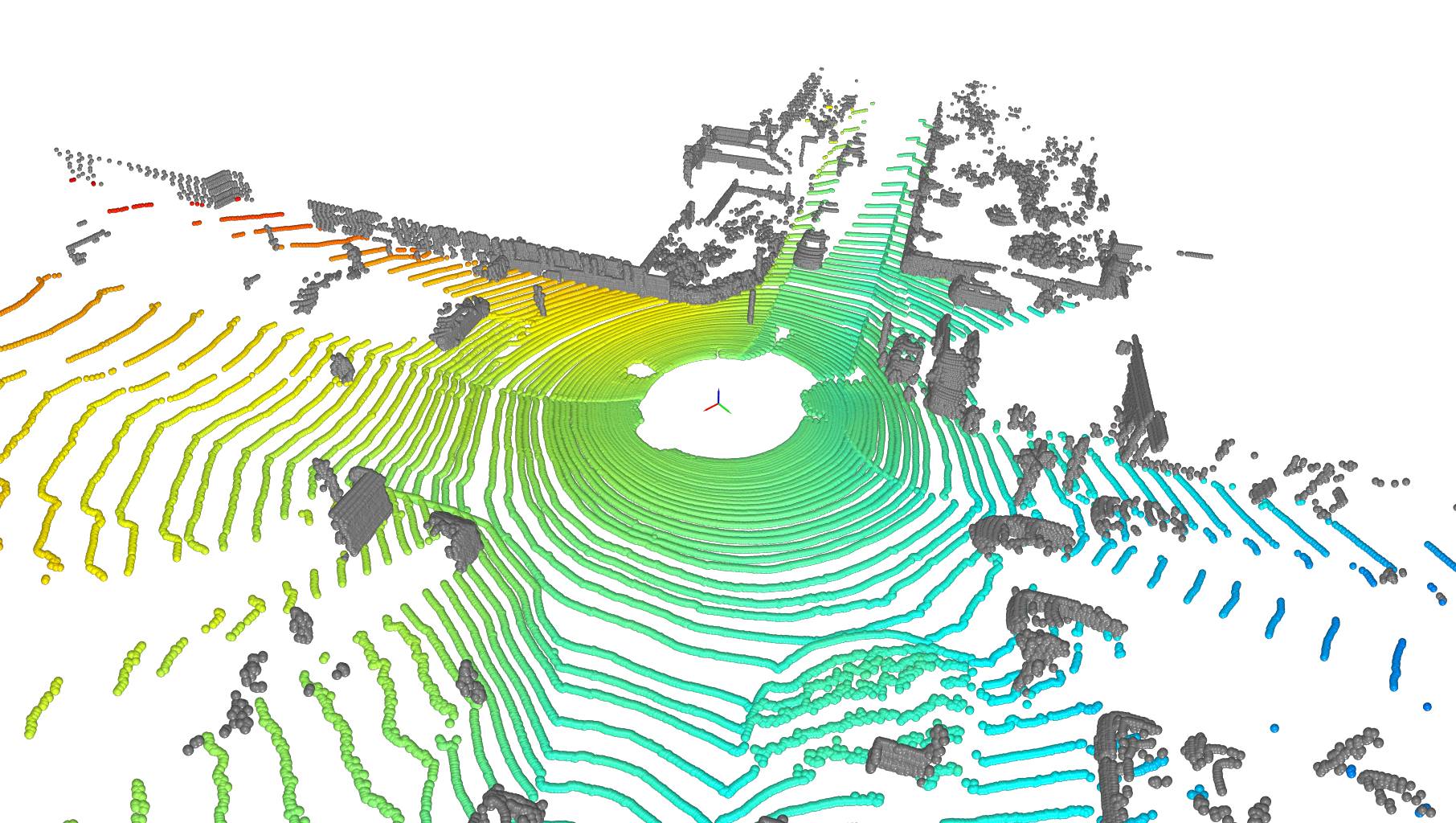}
    \includegraphics[width=0.32\linewidth]{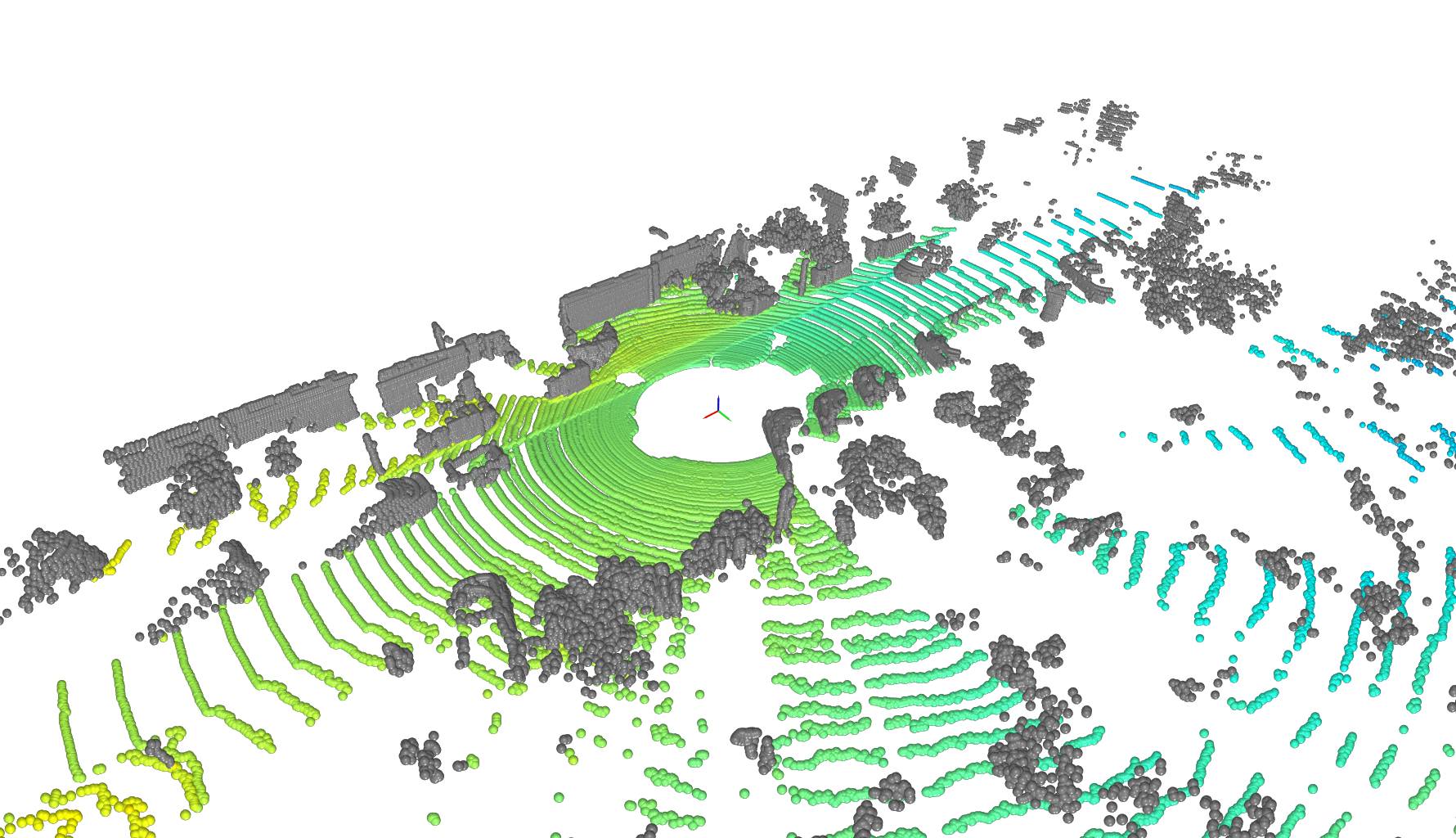}
    \includegraphics[width=0.32\linewidth]{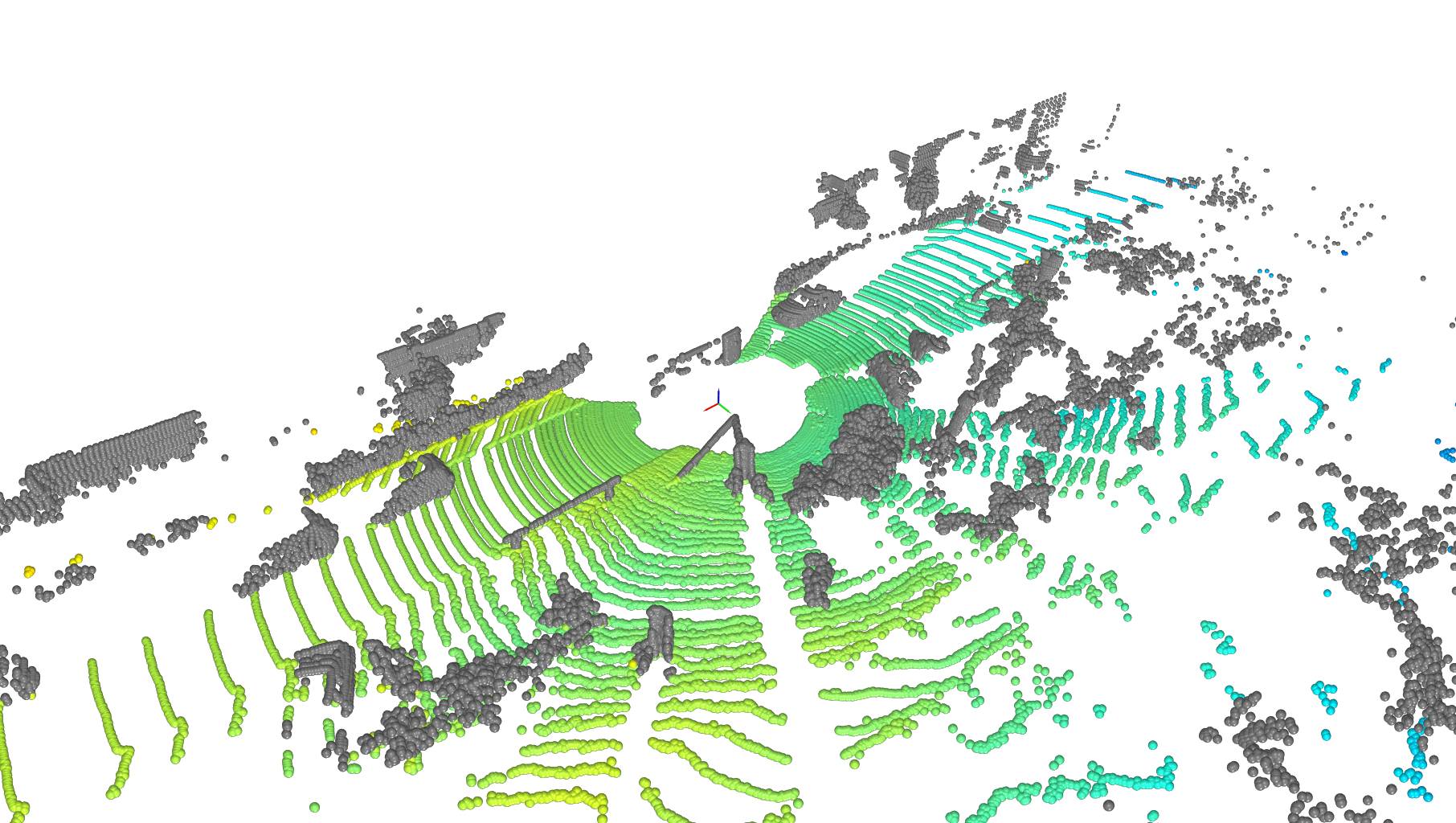}\\
    \vspace{2mm}
    \includegraphics[width=0.32\linewidth]{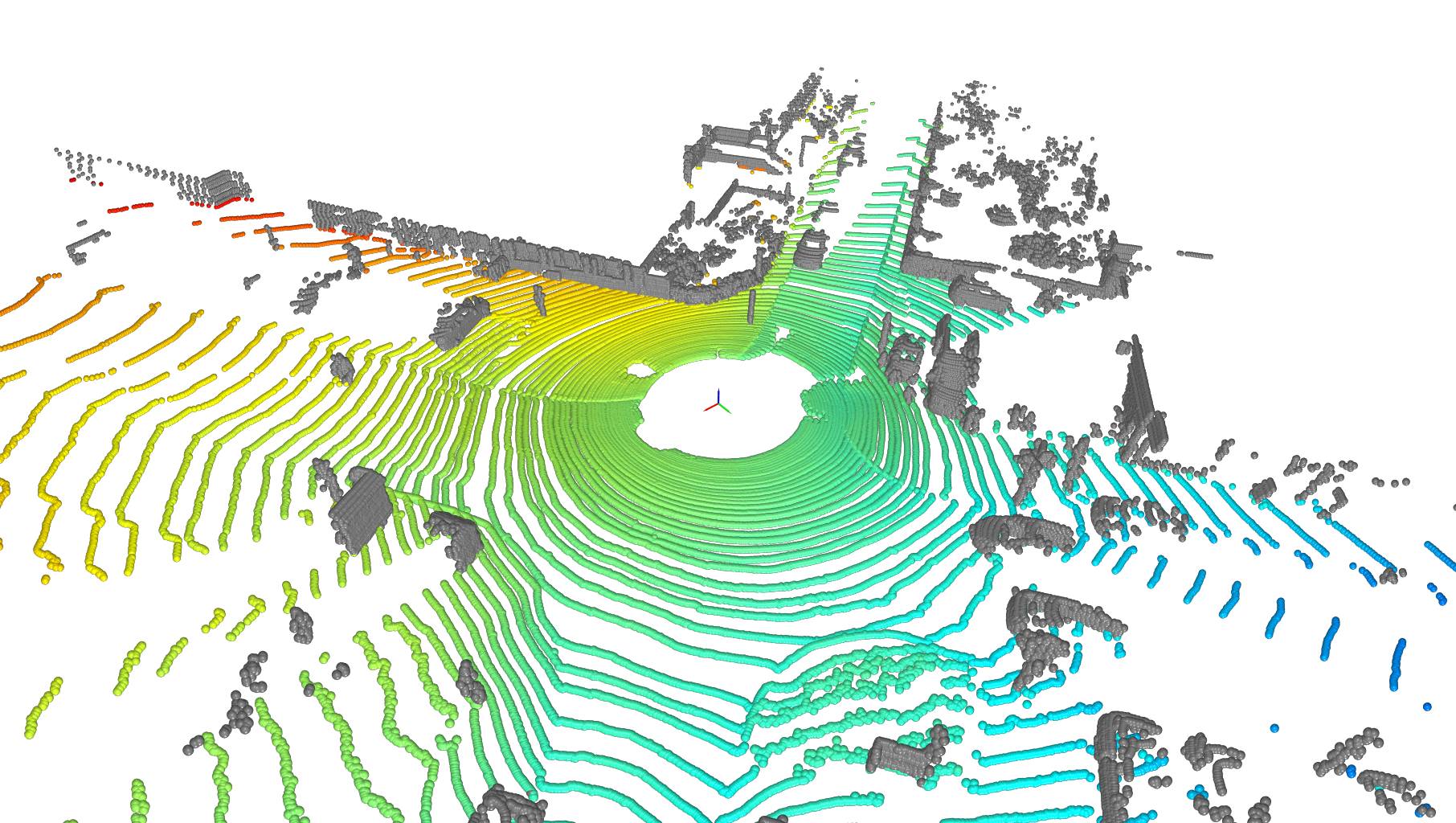}
    \includegraphics[width=0.32\linewidth]{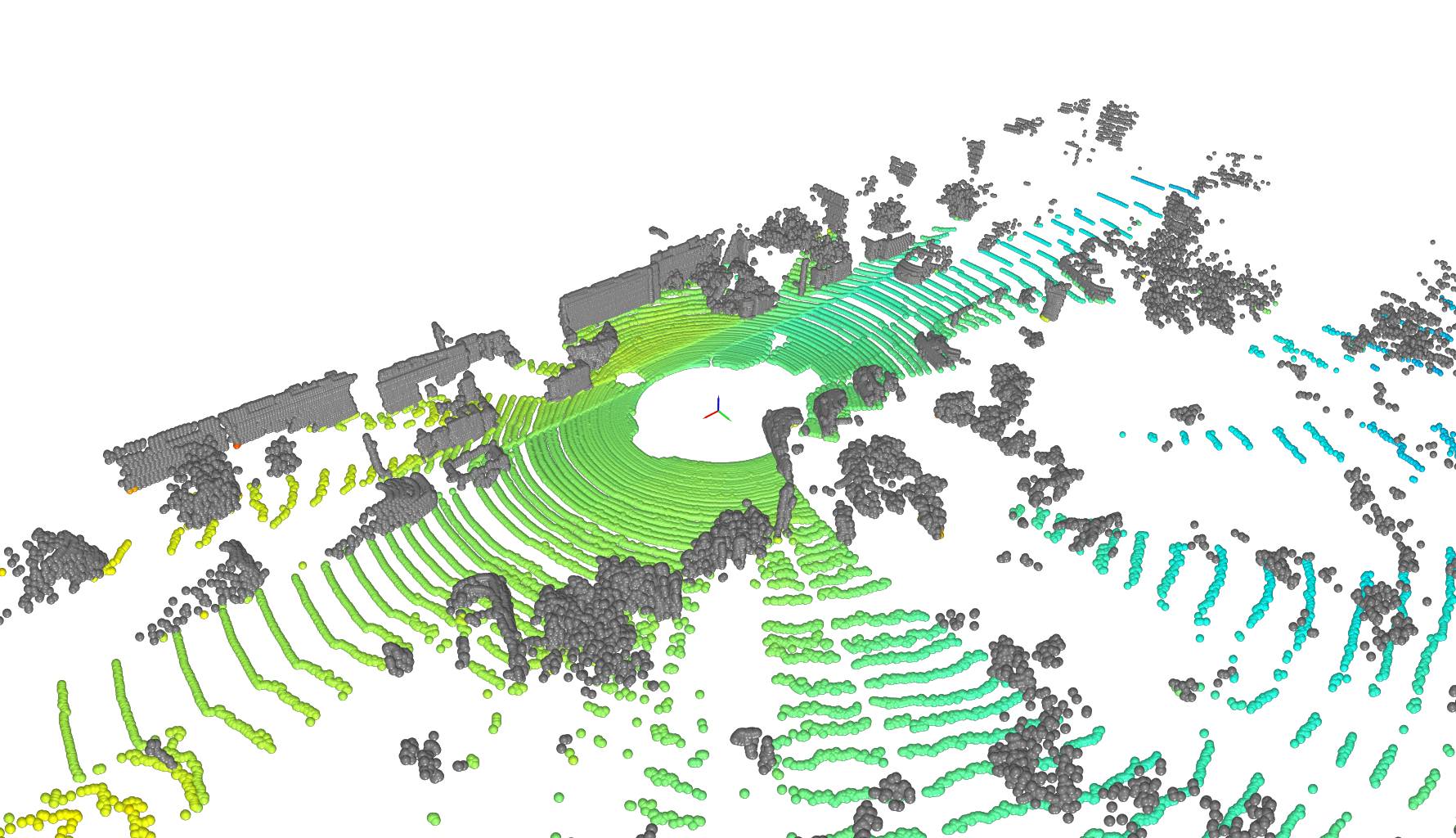}
    \includegraphics[width=0.32\linewidth]{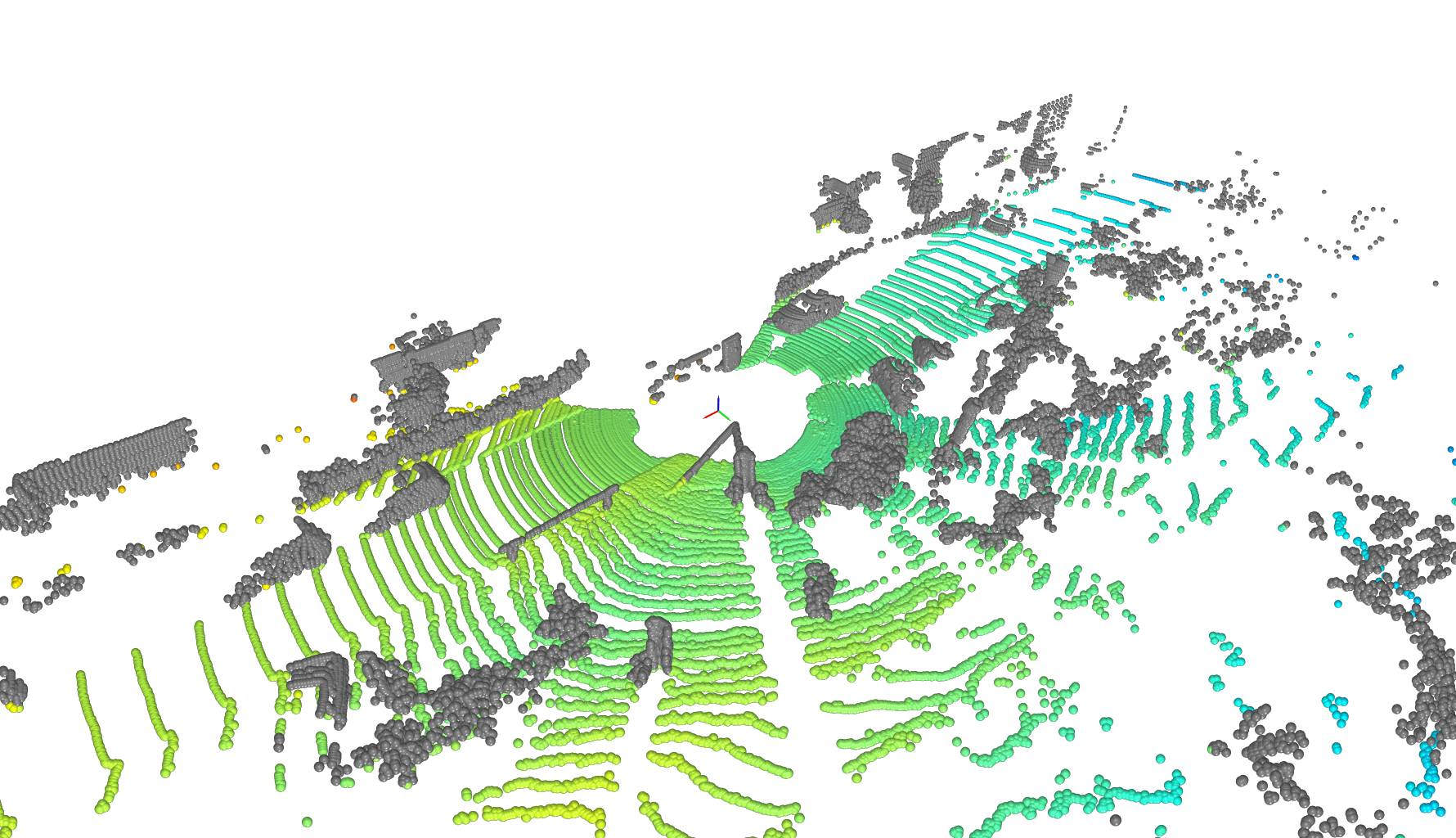}
    \caption{
    \textbf{Qualitative success cases on SemanticKITTI dataset~\cite{behley2019semantickitti}.} The top row shows ground truth labels, while the bottom row displays TerraSeg predictions. Ground points are color-coded by elevation; non-ground points are rendered in gray. From left to right, columns correspond to scan IDs \num{4006}, \num{2714}, and \num{2816}.}
    \label{fig:supp_success_semantickitti}
\end{figure*}

\begin{figure*}[h]
    \centering
    \includegraphics[width=0.32\linewidth]{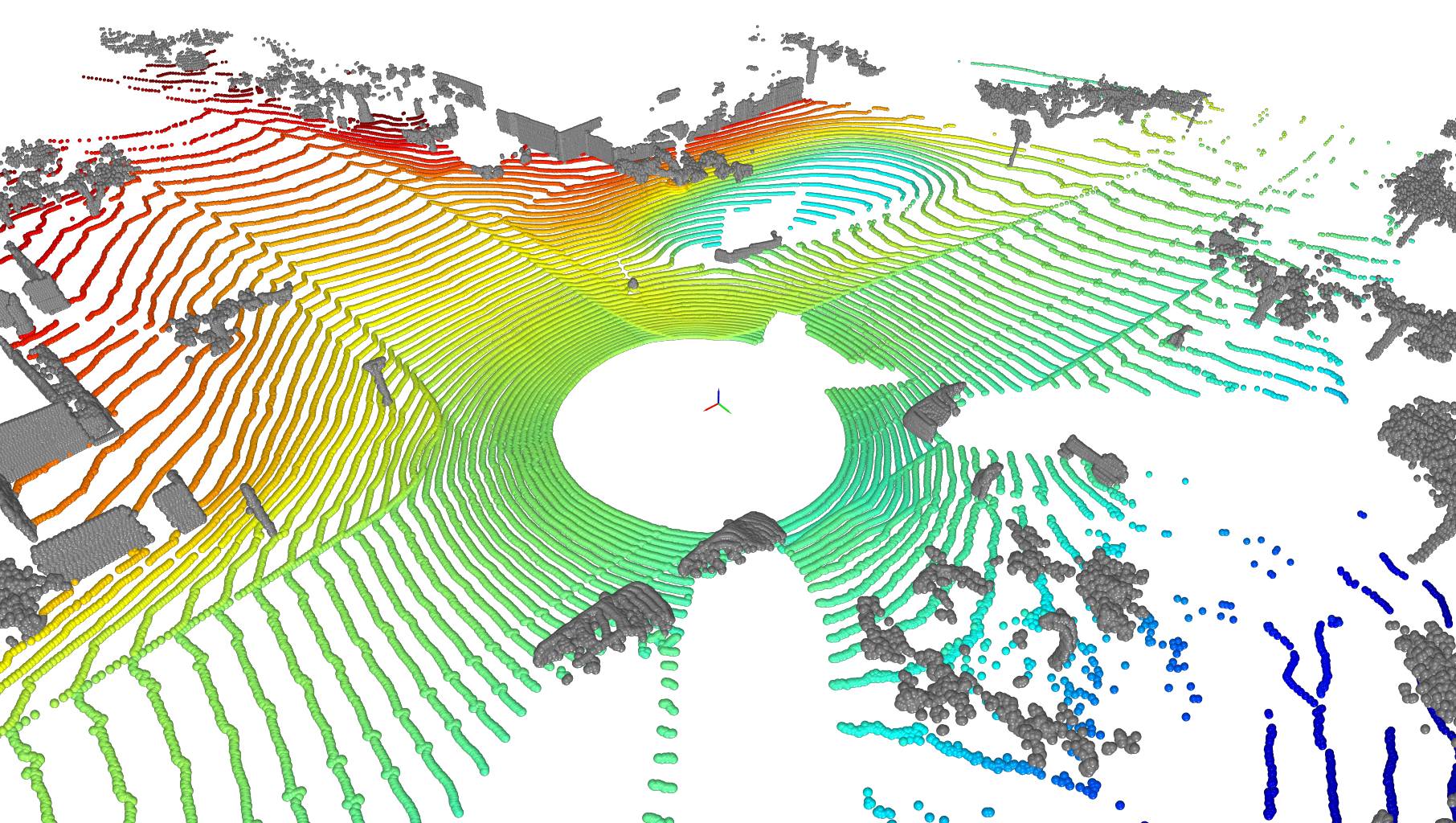}
    \includegraphics[width=0.32\linewidth]{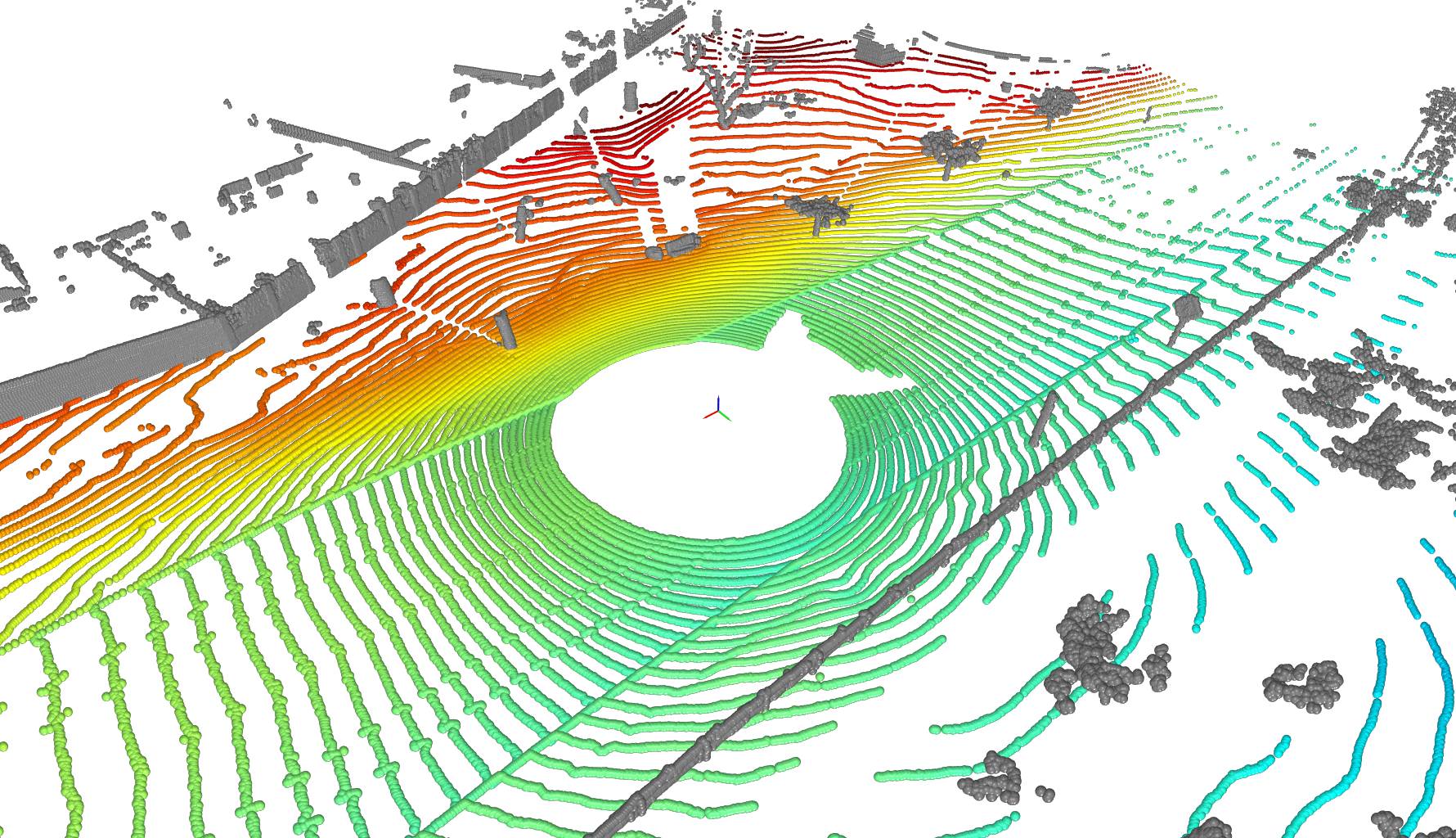}
    \includegraphics[width=0.32\linewidth]{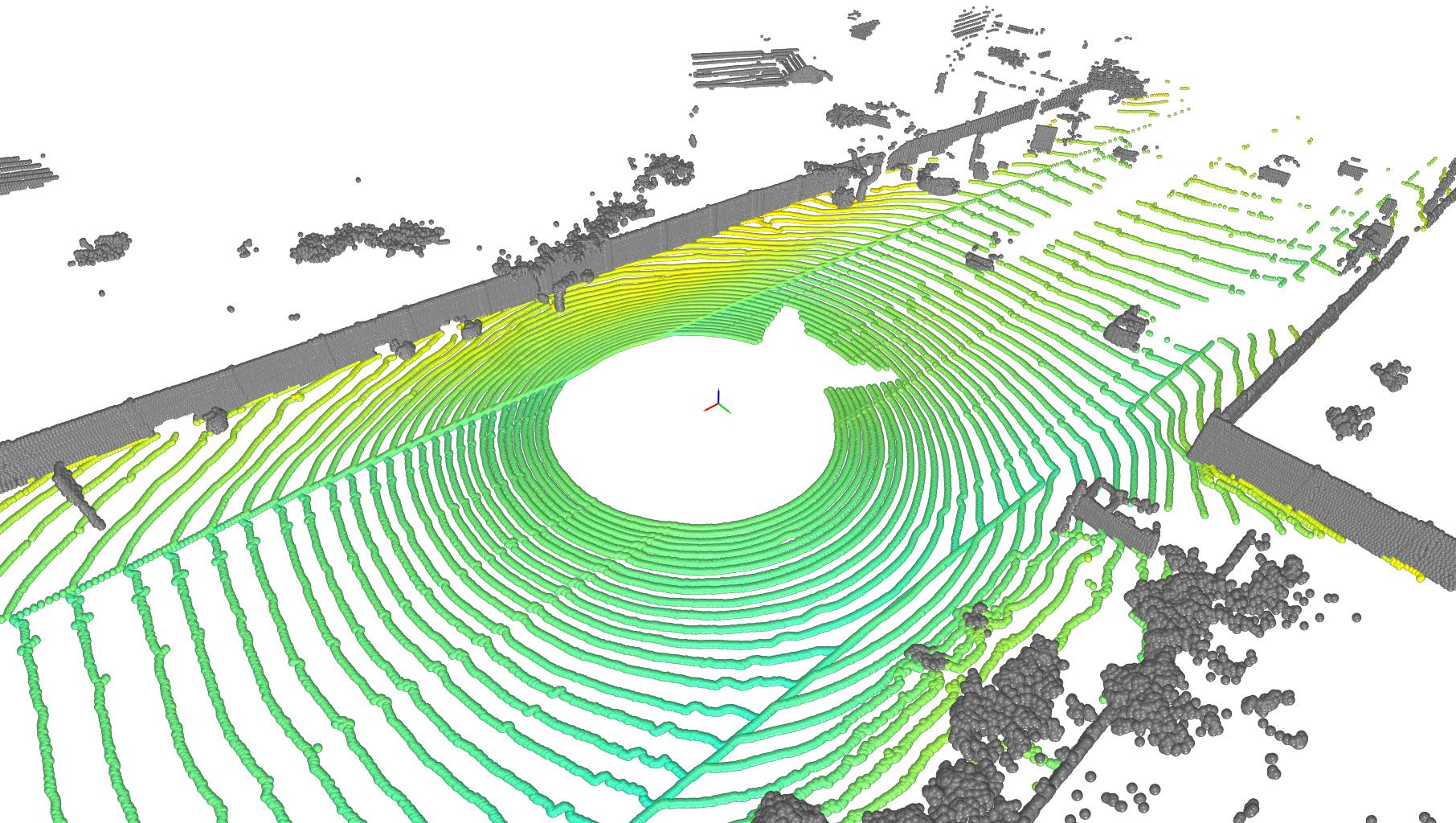}\\
    \vspace{2mm}
    \includegraphics[width=0.32\linewidth]{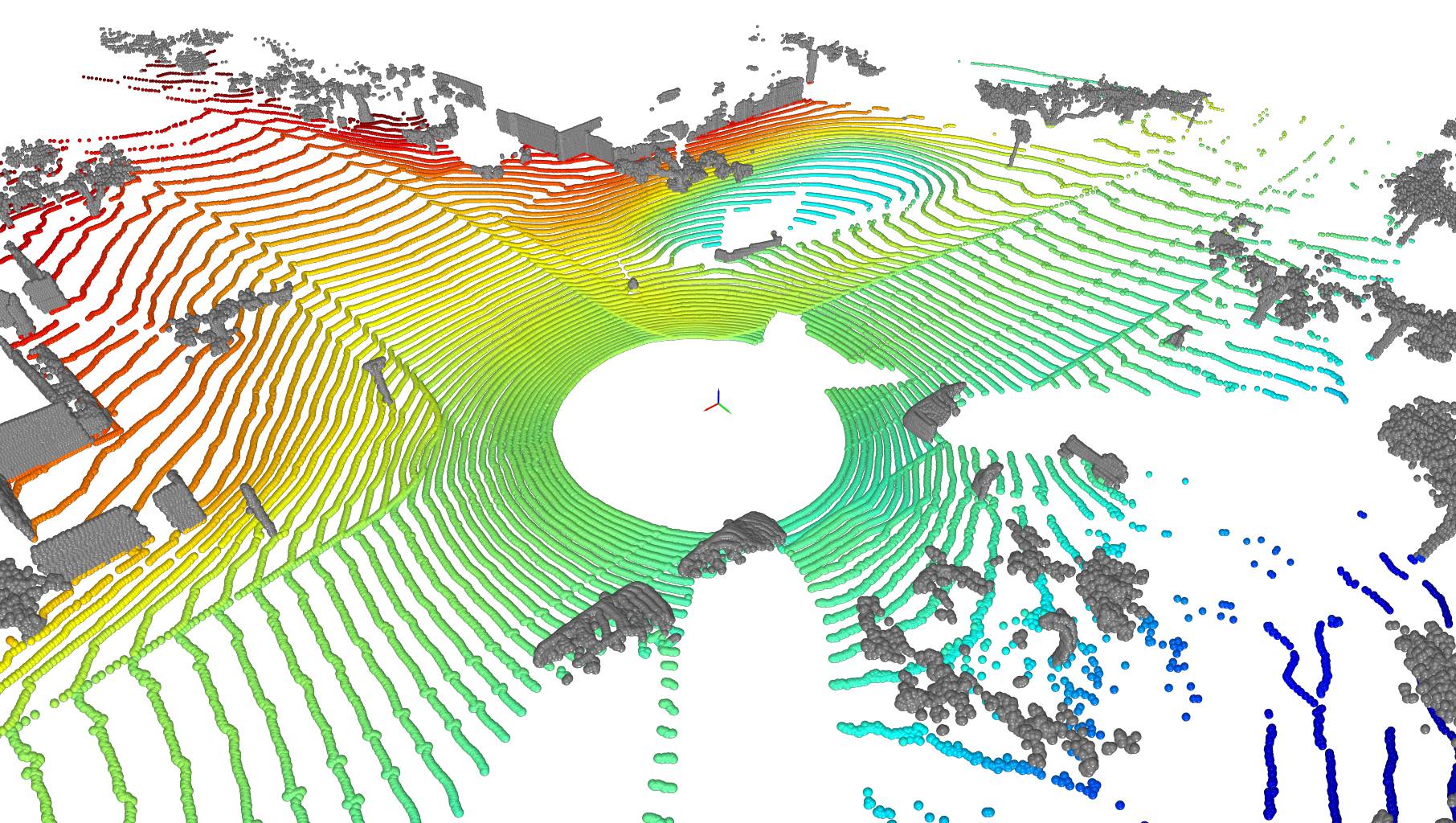}
    \includegraphics[width=0.32\linewidth]{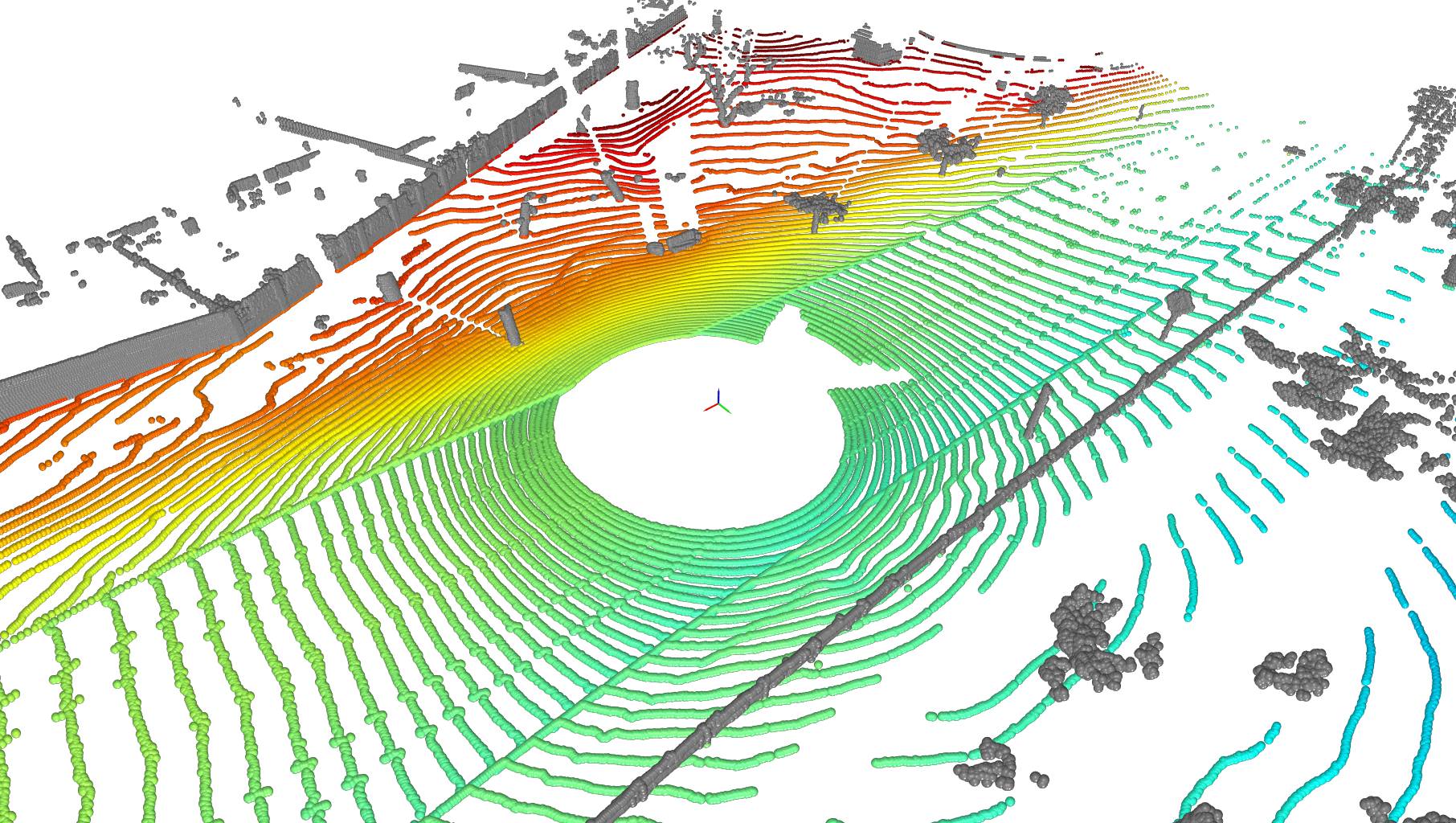}
    \includegraphics[width=0.32\linewidth]{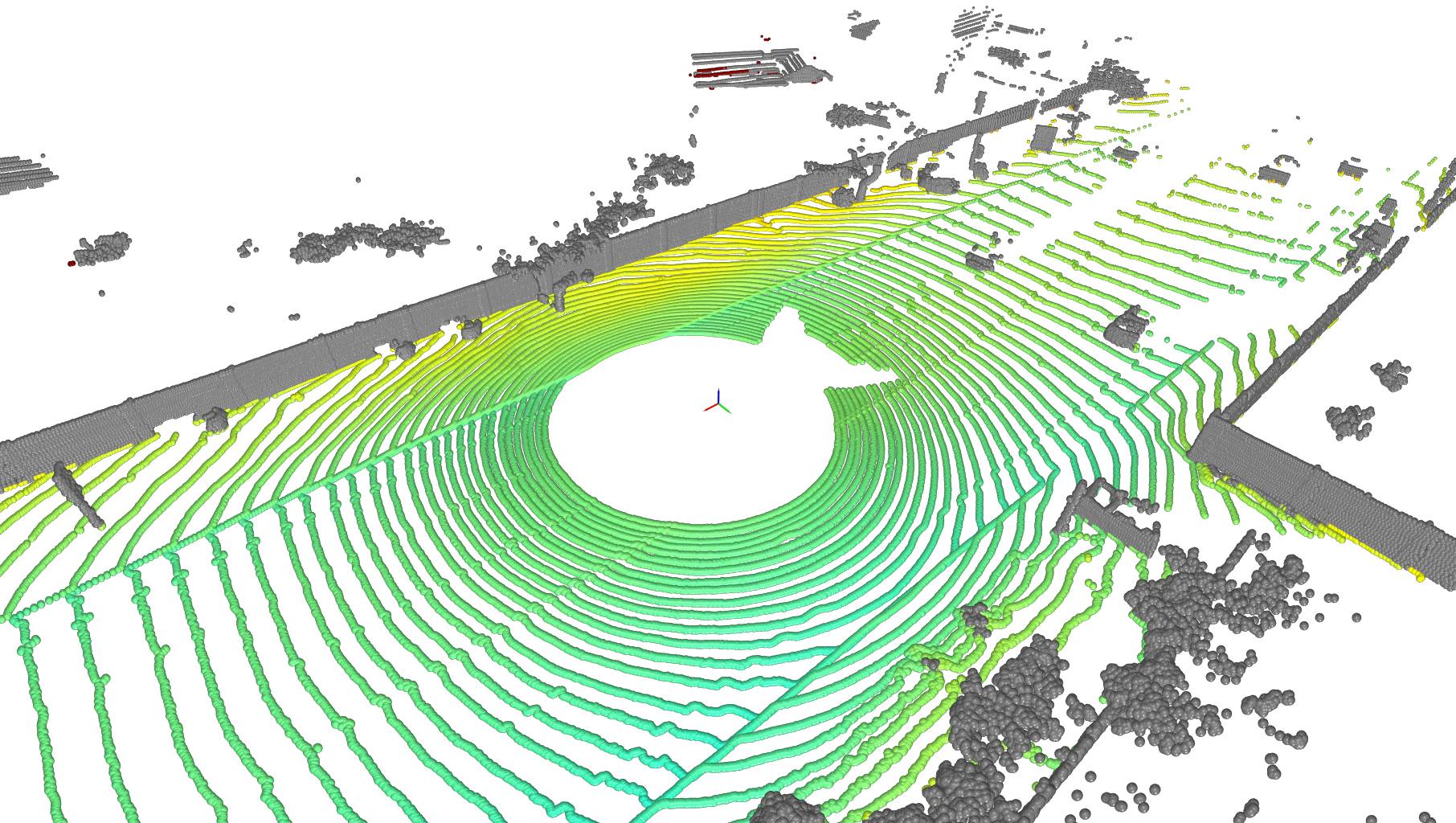}
    \caption{
    \textbf{Qualitative success cases on Waymo Perception dataset~\cite{sun2020scalability}.} The top row shows ground truth labels, while the bottom row displays TerraSeg predictions. Ground points are color-coded by elevation; non-ground points are rendered in gray. From left to right, columns correspond to scan IDs \num{3659}, \num{3045}, and \num{3505}.}
    \label{fig:supp_success_waymo}
\end{figure*}

In this section, we present qualitative success cases of TerraSeg-B.
Figures \ref{fig:supp_success_nuscenes}, \ref{fig:supp_success_semantickitti}, and \ref{fig:supp_success_waymo} illustrate the segmentation performance on the nuScenes~\cite{caesar2020nuscenes}, SemanticKITTI~\cite{behley2019semantickitti}, and Waymo Perception~\cite{sun2020scalability} validation splits, respectively.
Each figure is arranged in two rows: the top row displays the ground truth annotations, while the bottom row shows the predictions generated by our domain-agnostic TerraSeg model.
These results demonstrate TerraSeg's robustness in delineating complex ground geometries and adapting to varying terrain flatness across vastly different sensor configurations without relying on manual annotations.

\subsection{Failure cases}

\begin{figure*}[h]
    \centering
    \includegraphics[width=0.32\linewidth]{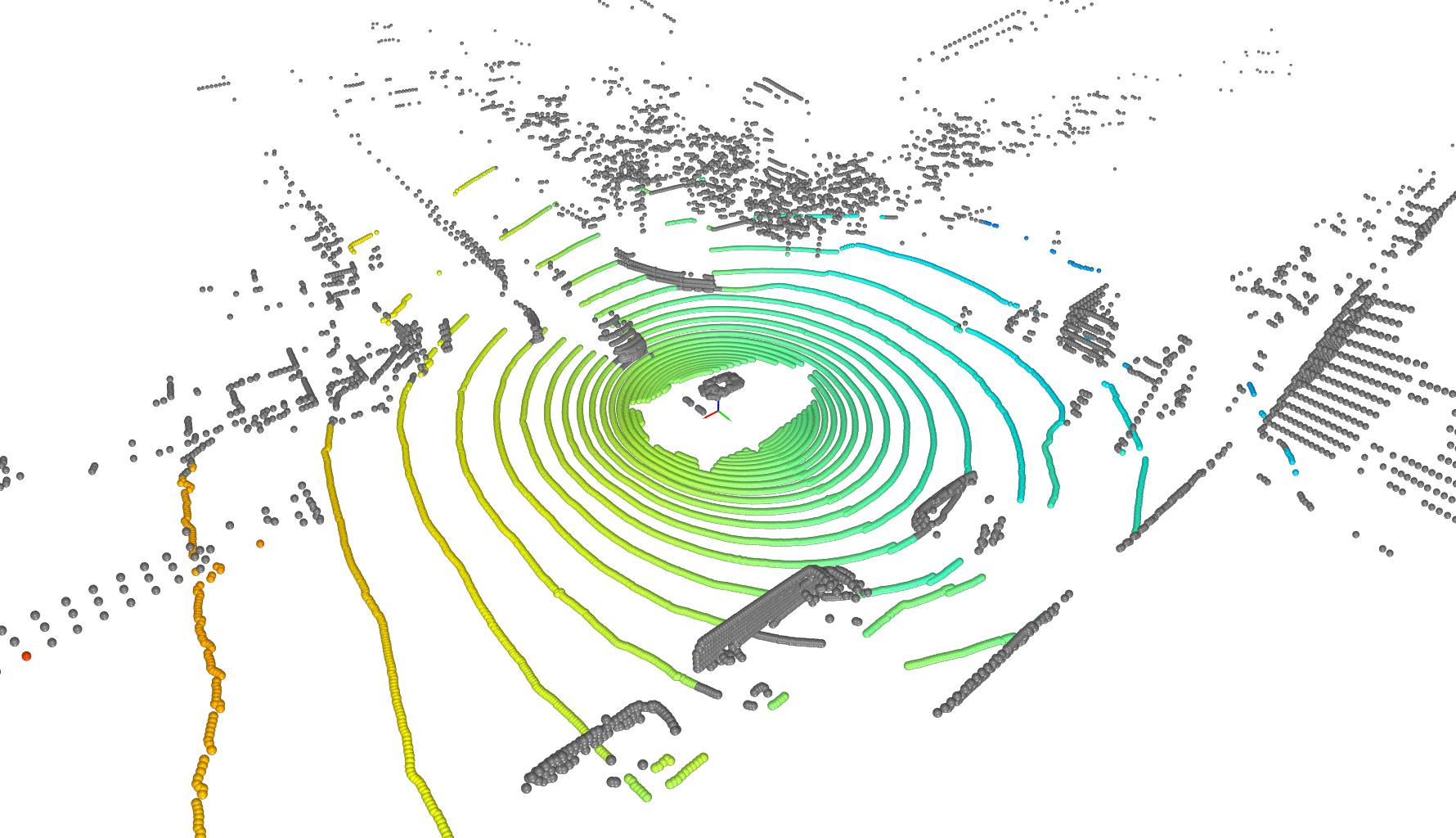}
    \includegraphics[width=0.32\linewidth]{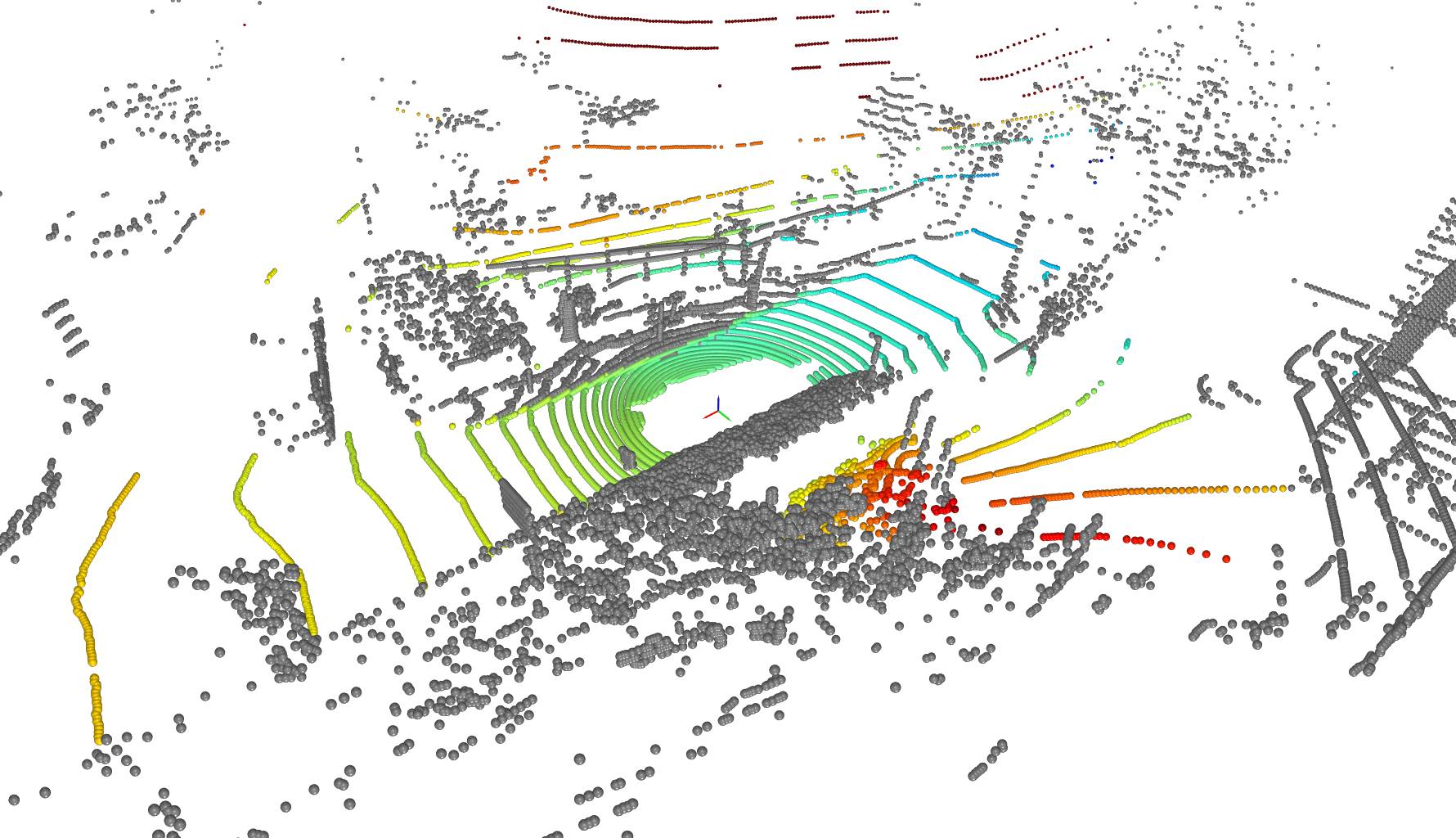}
    \includegraphics[width=0.32\linewidth]{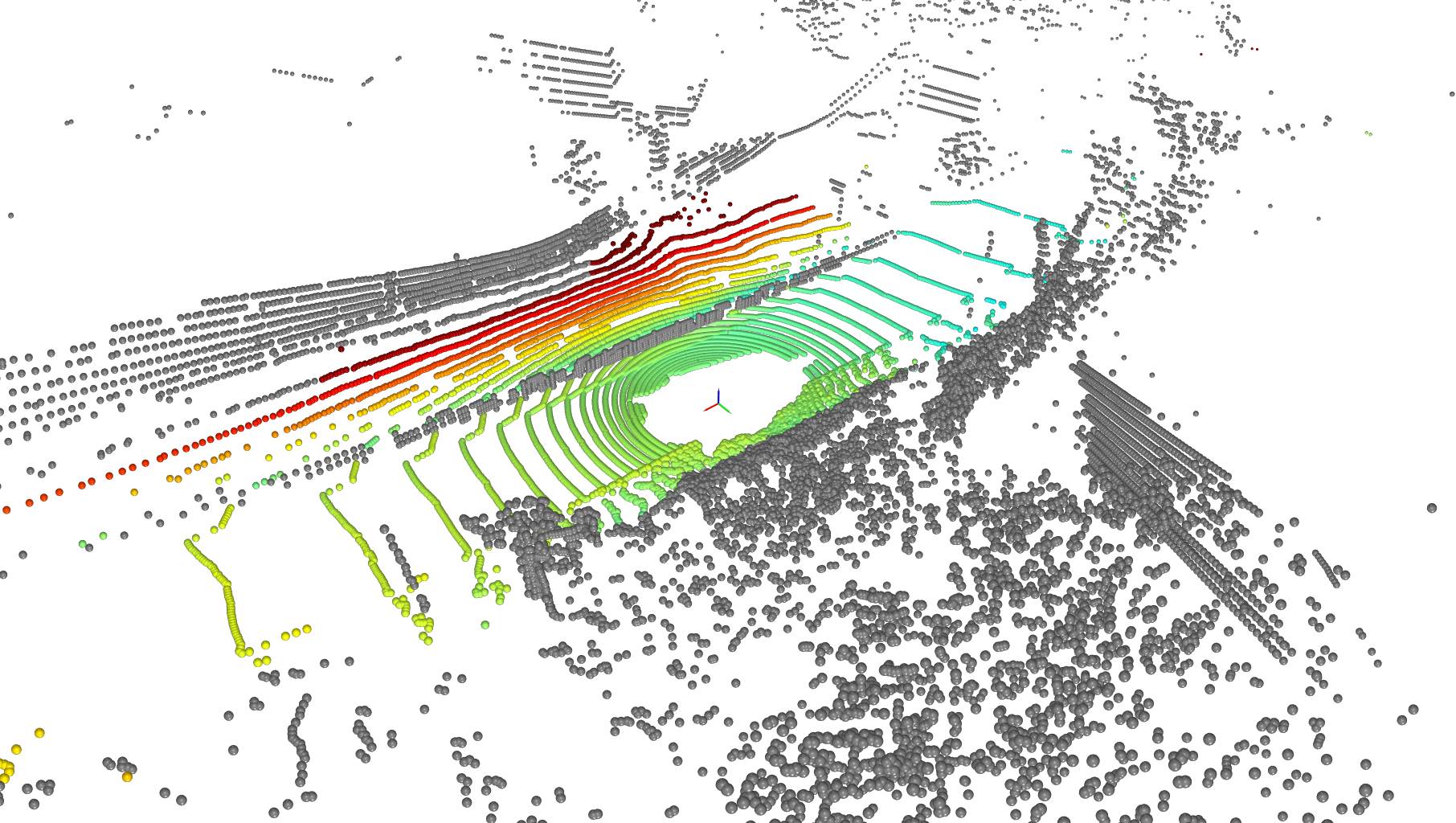}\\
    \vspace{2mm}
    \includegraphics[width=0.32\linewidth]{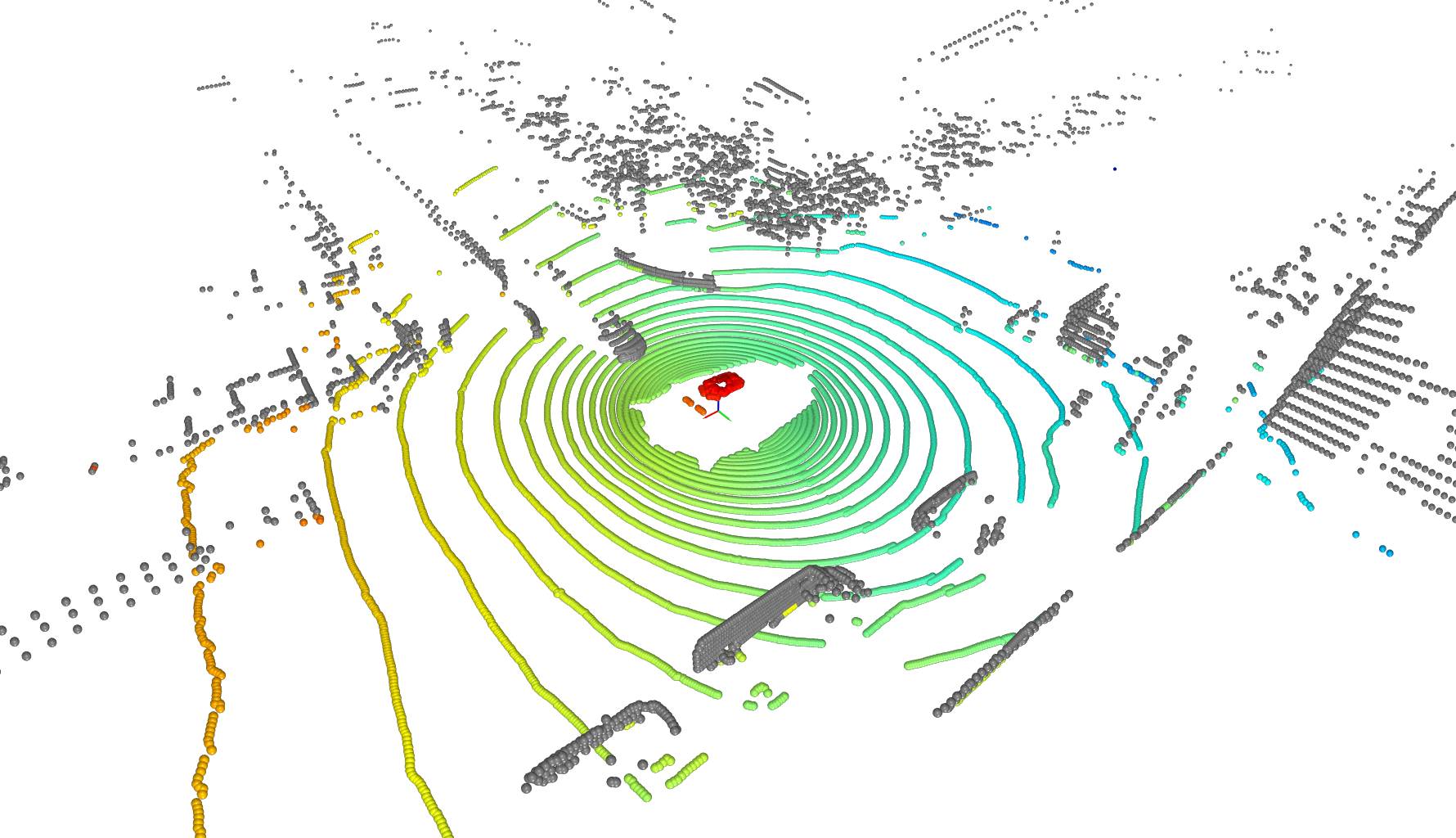}
    \includegraphics[width=0.32\linewidth]{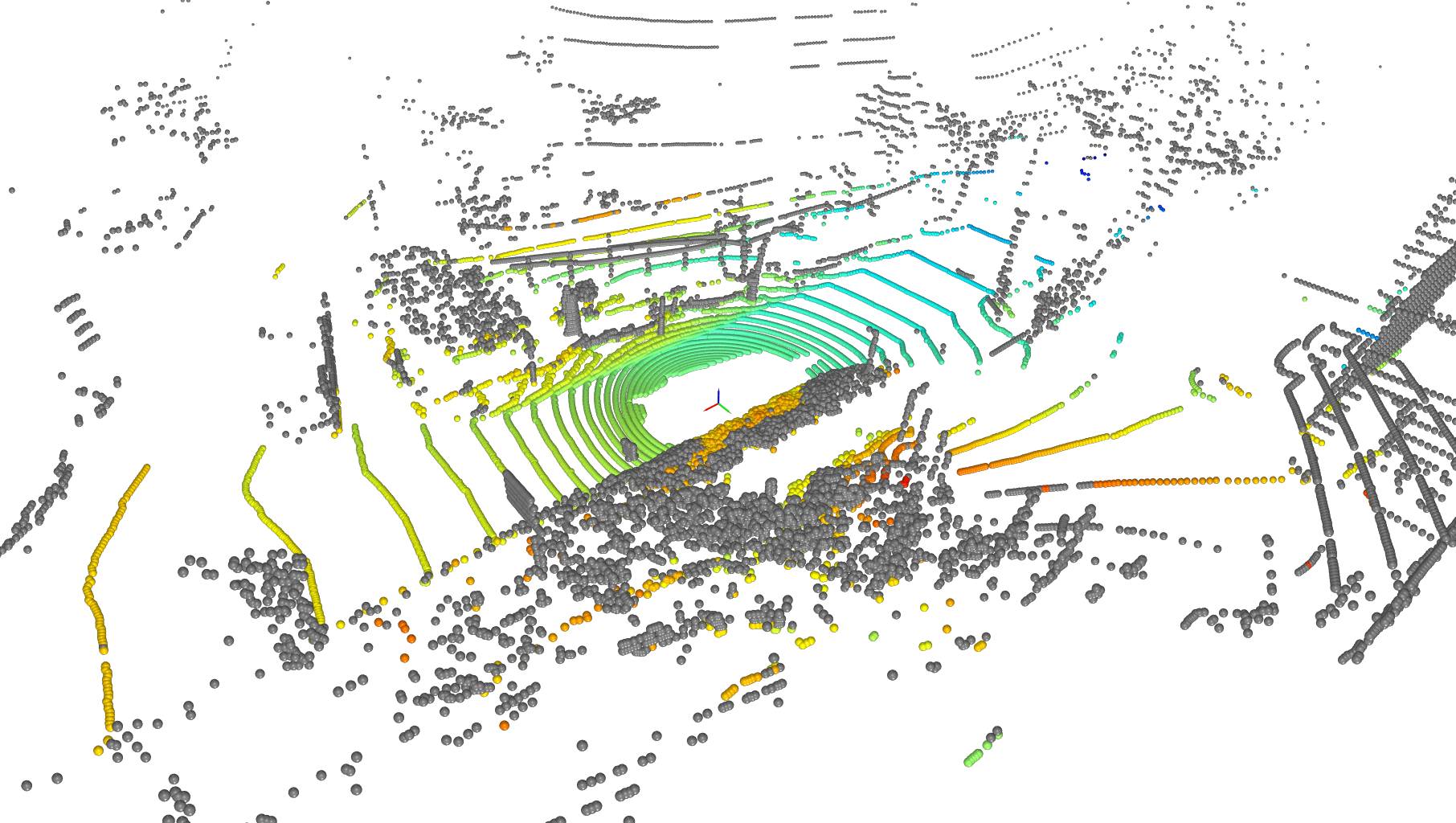}
    \includegraphics[width=0.32\linewidth]{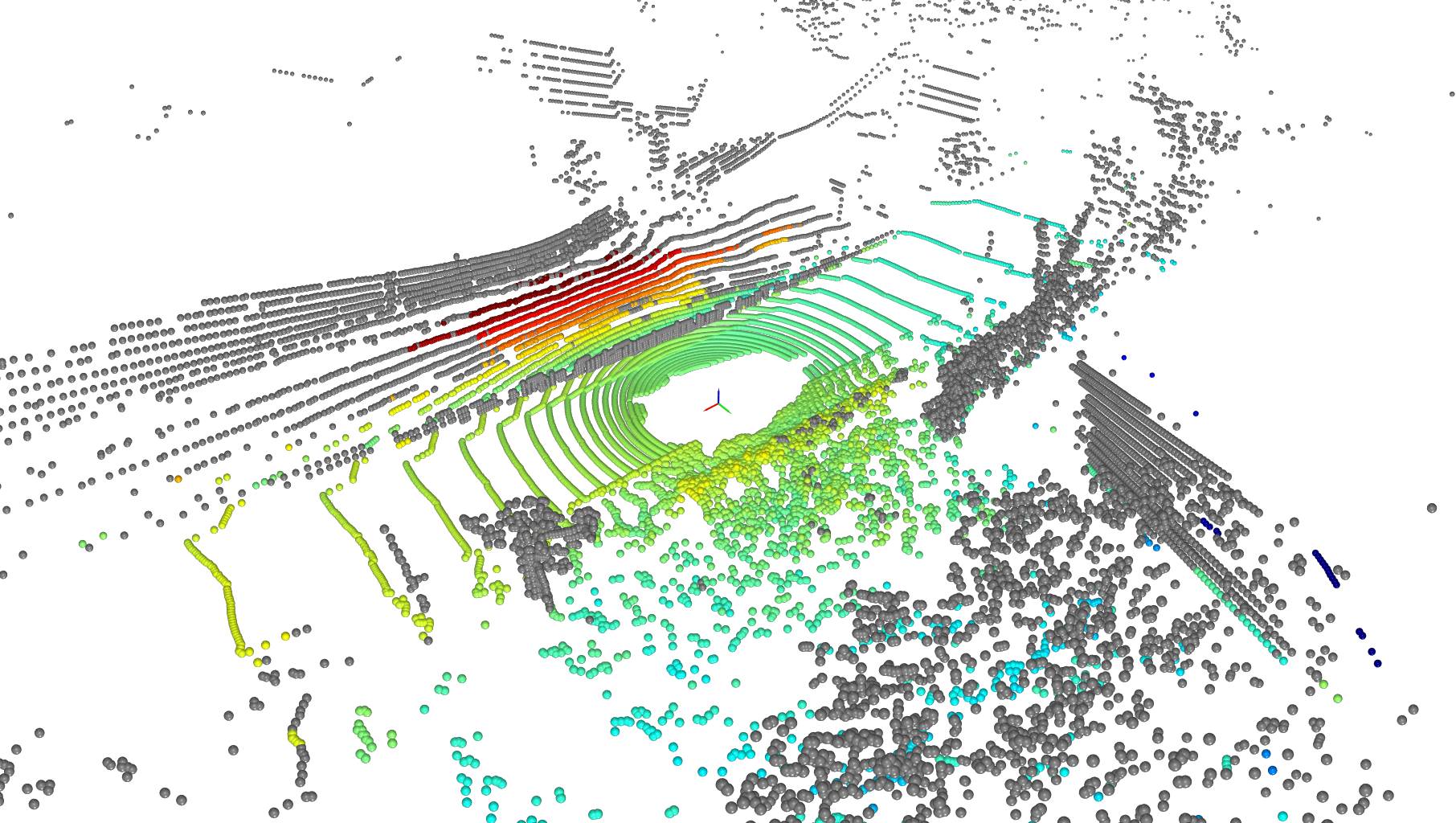}
    \caption{
    \textbf{Qualitative failure cases on nuScenes dataset~\cite{caesar2020nuscenes}.} The top row shows ground truth labels, while the bottom row displays TerraSeg predictions. Ground points are color-coded by elevation; non-ground points are rendered in gray. From left to right, columns correspond to scan IDs \num{2124}, \num{4848}, and \num{62}.}
    \label{fig:supp_fail_nuscenes}
\end{figure*}

\begin{figure*}[h]
    \centering
    \includegraphics[width=0.32\linewidth]{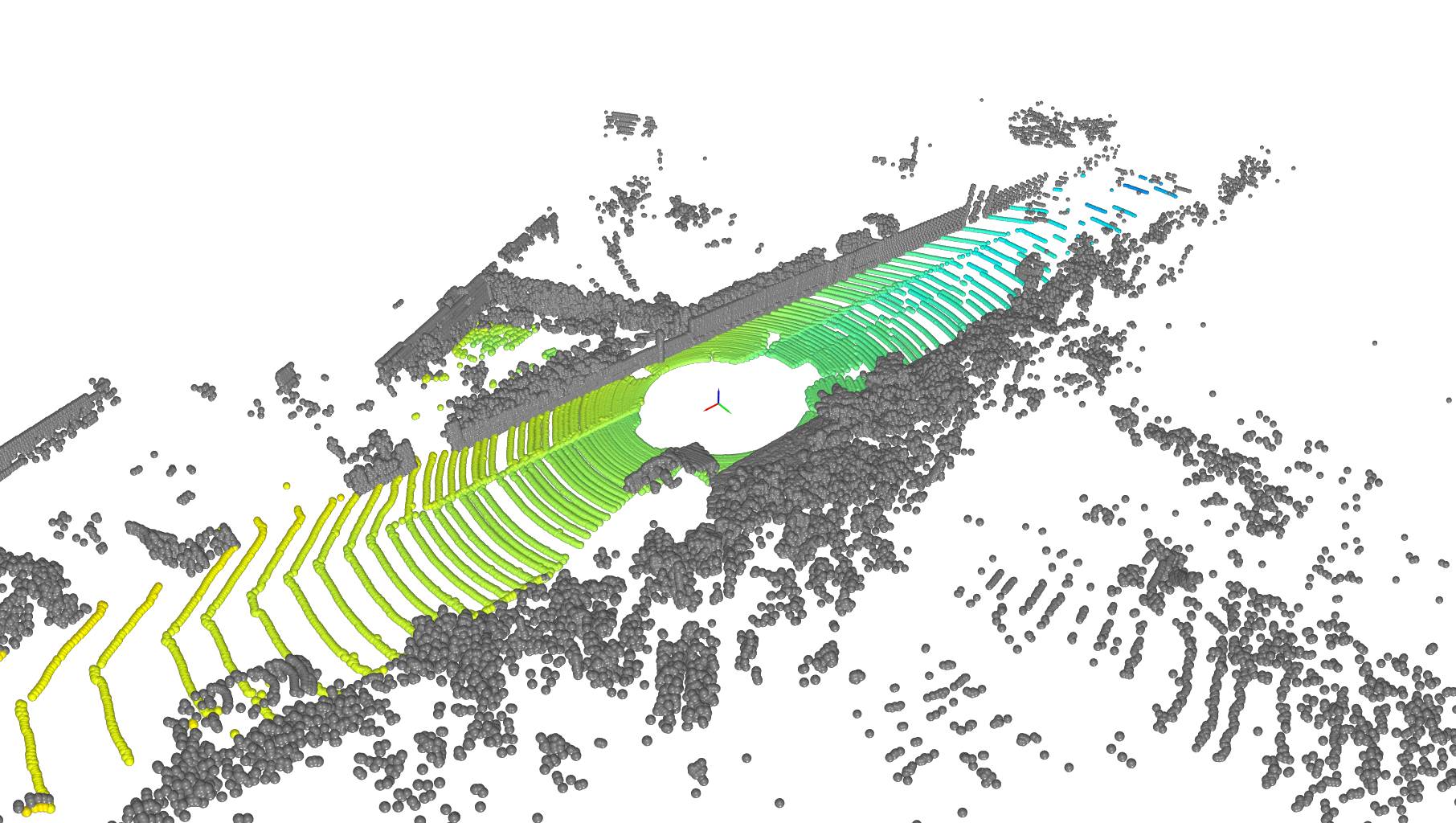}
    \includegraphics[width=0.32\linewidth]{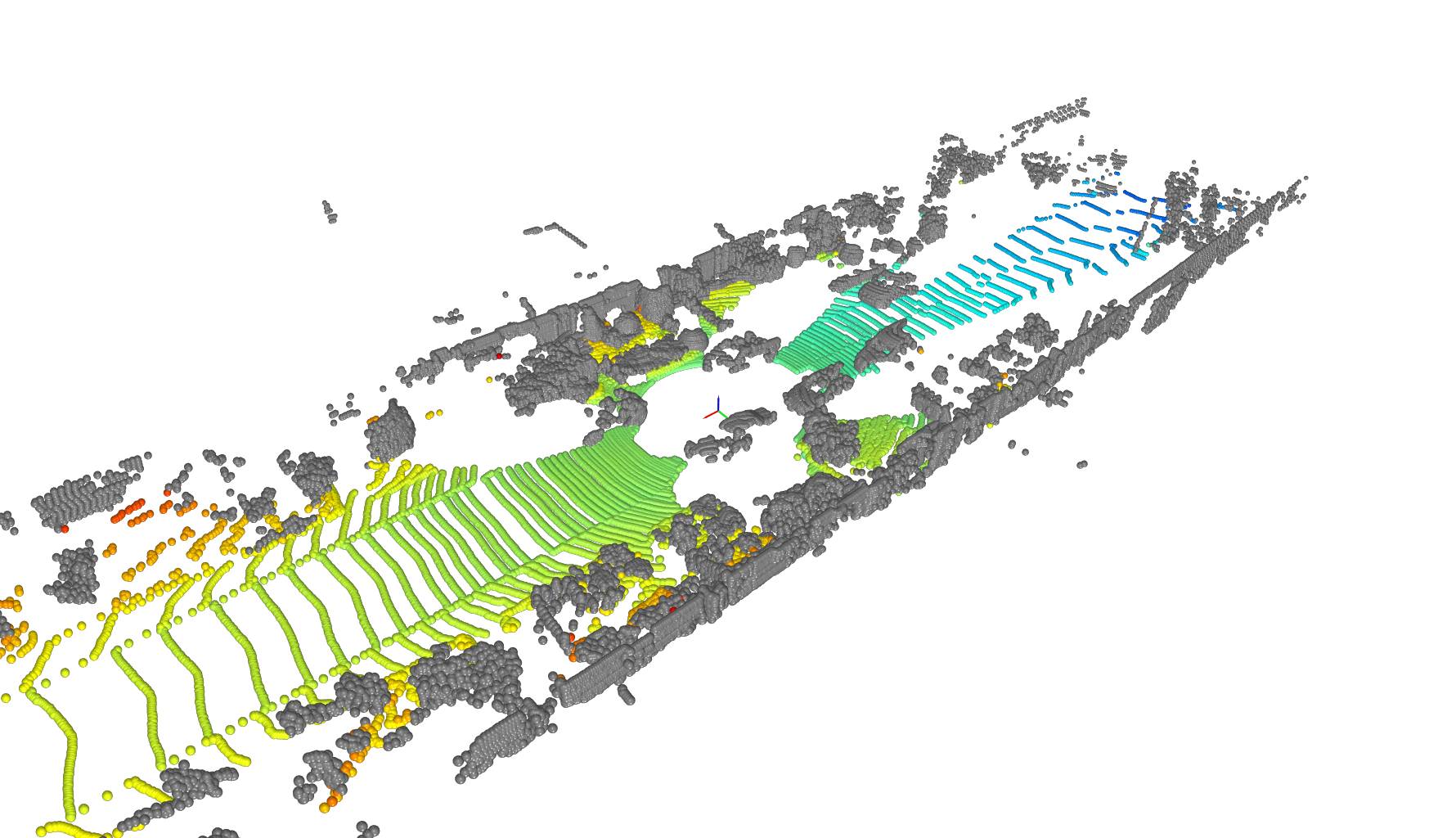}
    \includegraphics[width=0.32\linewidth]{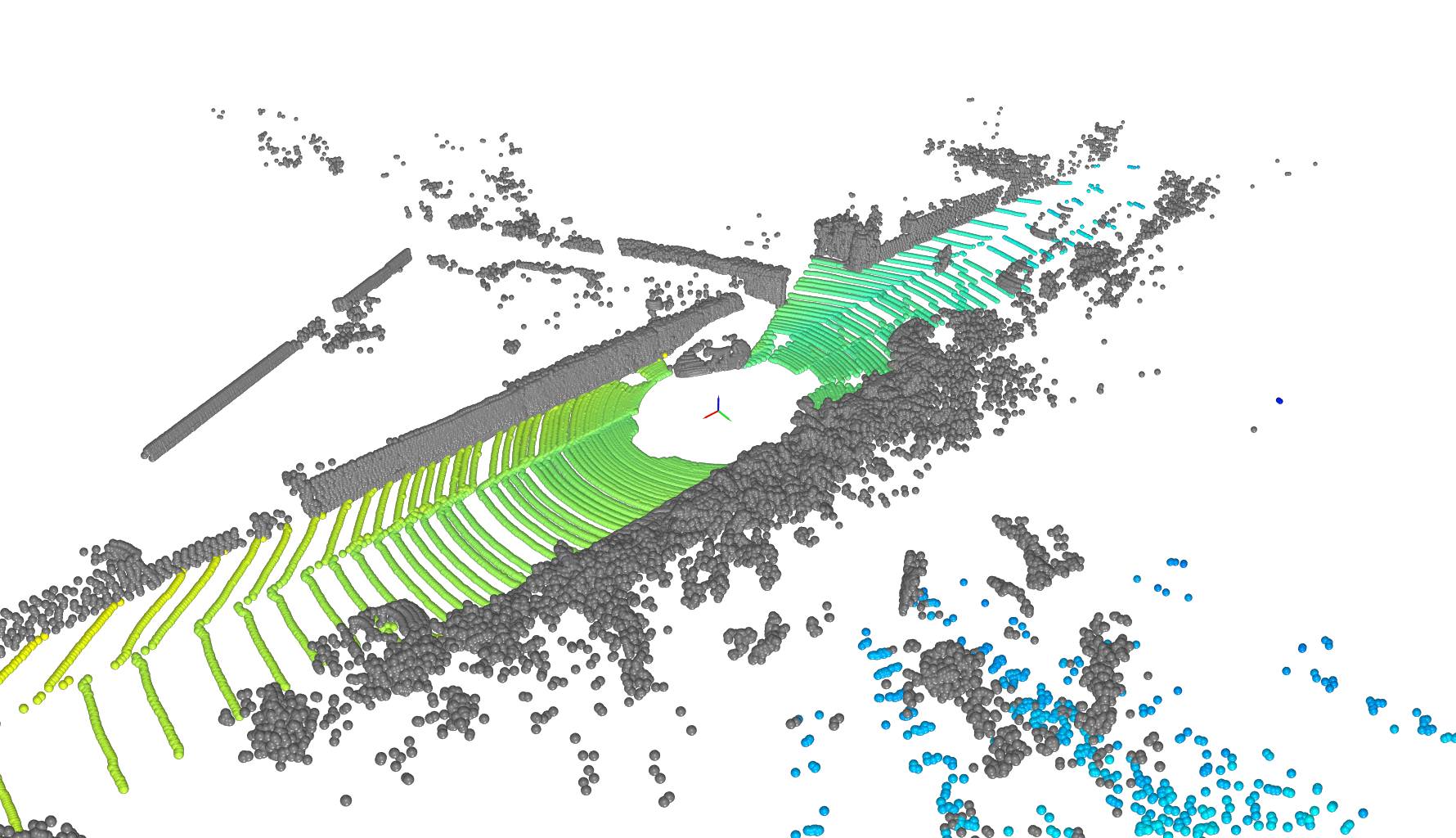}\\
    \vspace{2mm}
    \includegraphics[width=0.32\linewidth]{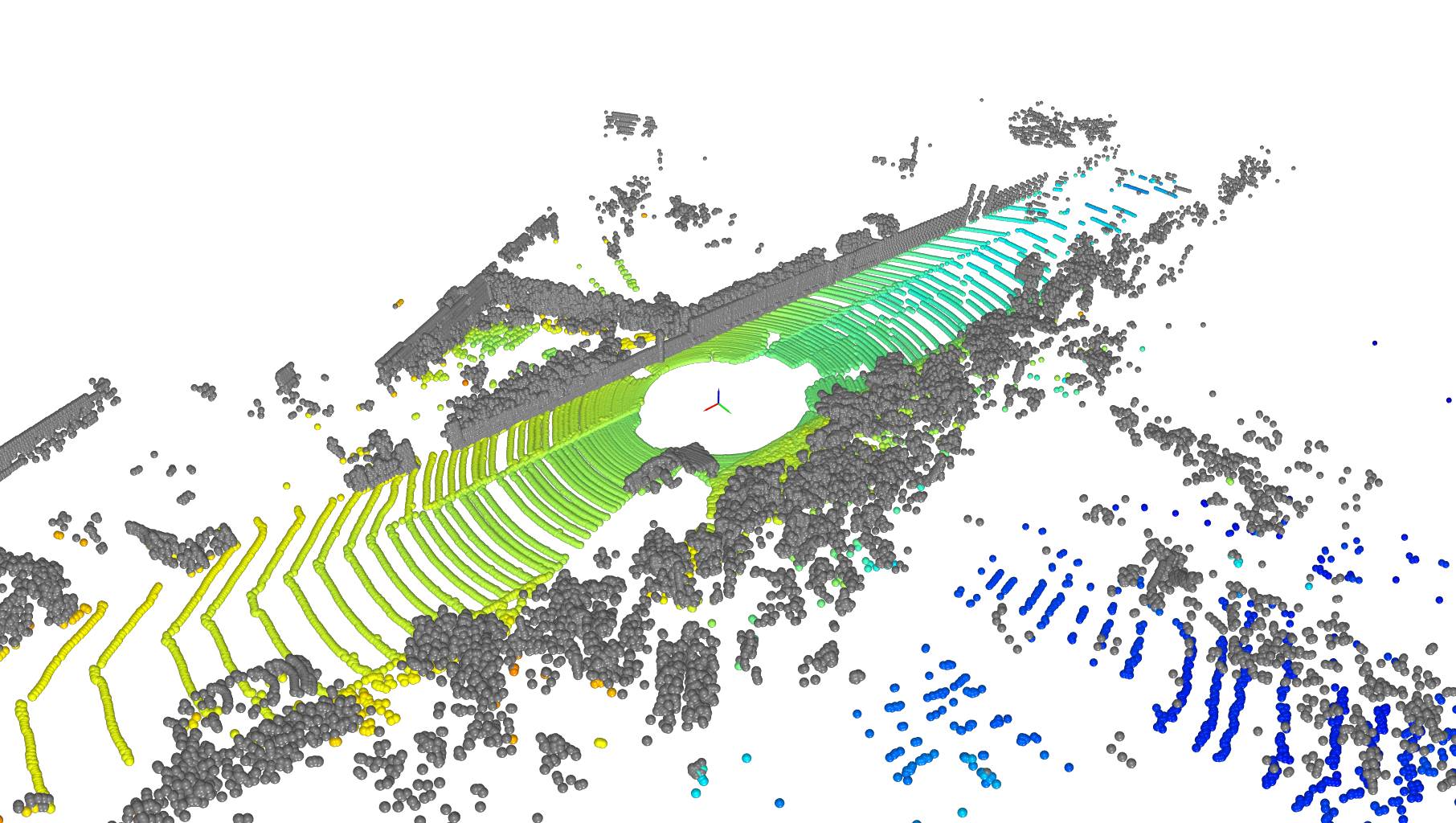}
    \includegraphics[width=0.32\linewidth]{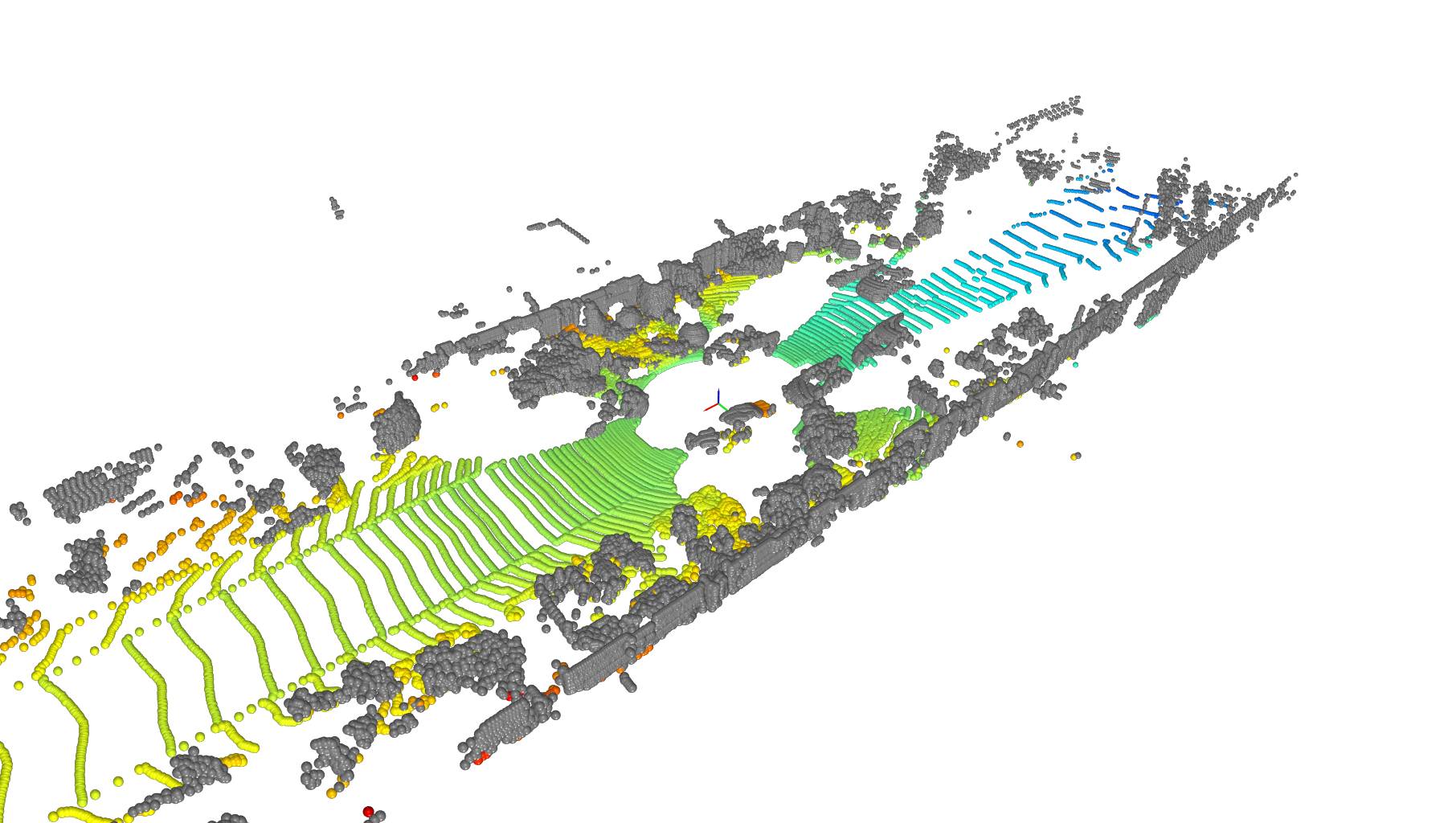}
    \includegraphics[width=0.32\linewidth]{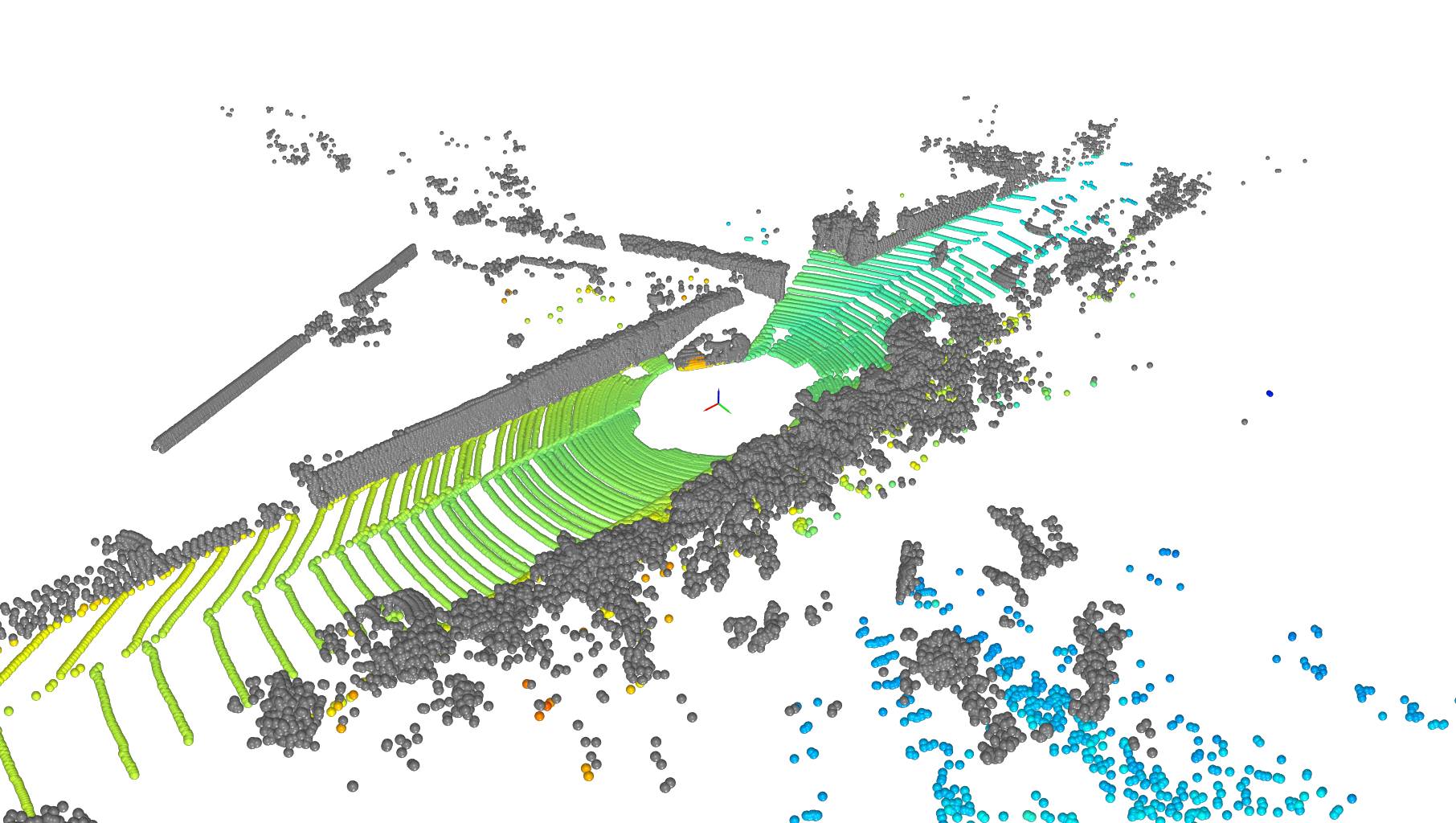}
    \caption{
    \textbf{Qualitative failure cases on SemanticKITTI dataset~\cite{behley2019semantickitti}.} The top row shows ground truth labels, while the bottom row displays TerraSeg predictions. Ground points are color-coded by elevation; non-ground points are rendered in gray. From left to right, columns correspond to scan IDs \num{3300}, \num{947}, and \num{3173}.}
    \label{fig:supp_fail_semantickitti}
\end{figure*}

\begin{figure*}[h]
    \centering
    \includegraphics[width=0.32\linewidth]{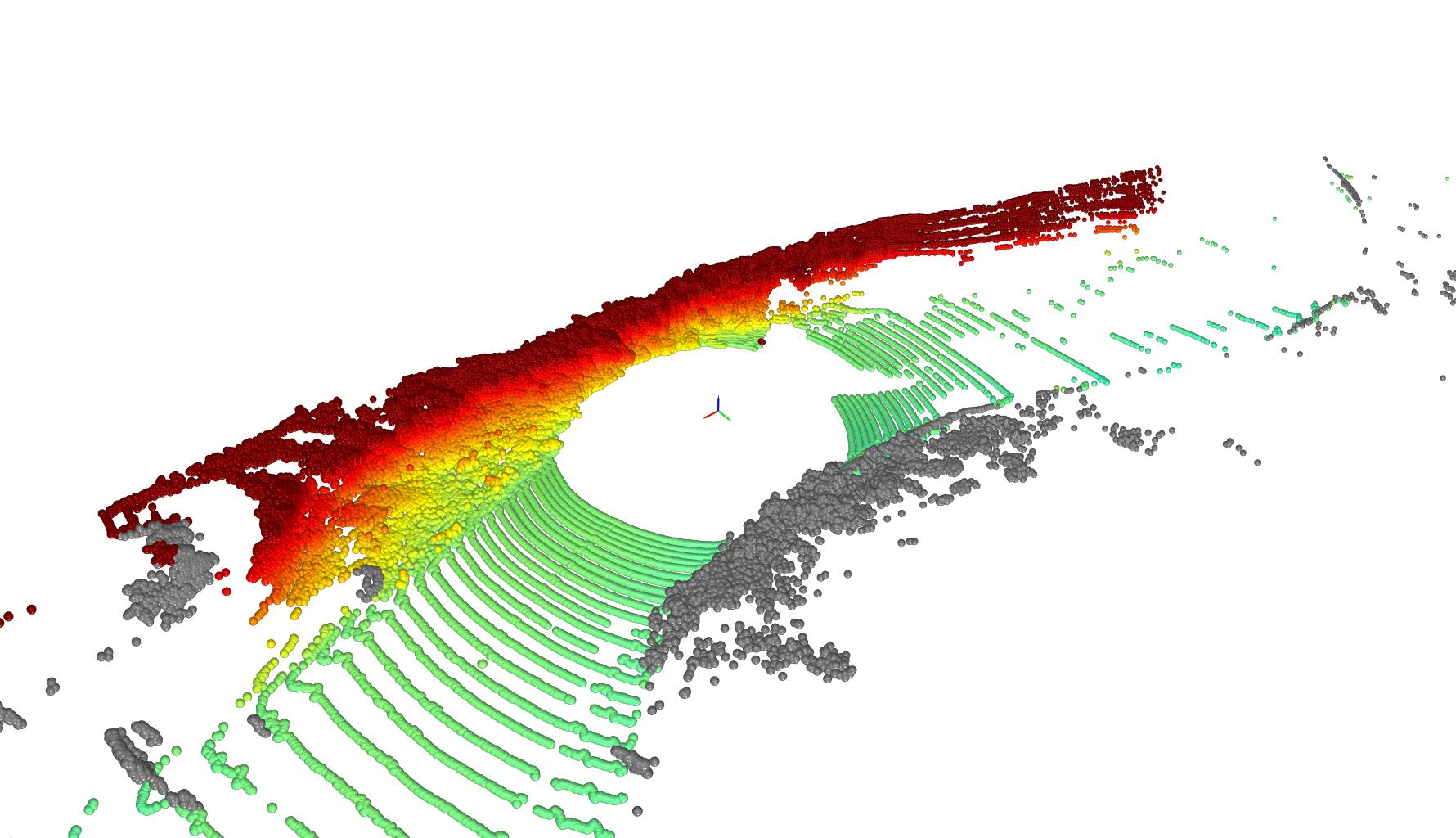}
    \includegraphics[width=0.32\linewidth]{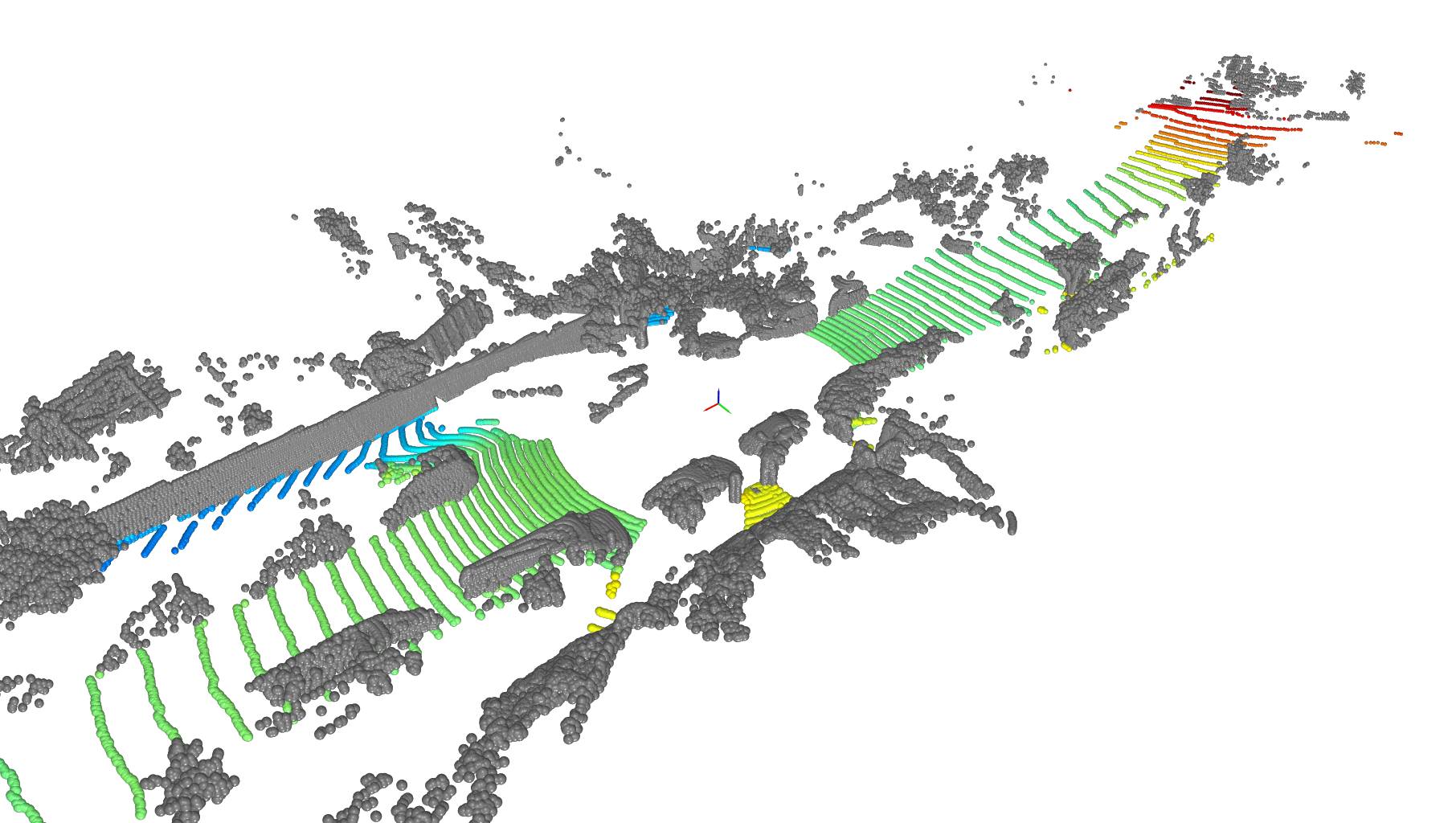}
    \includegraphics[width=0.32\linewidth]{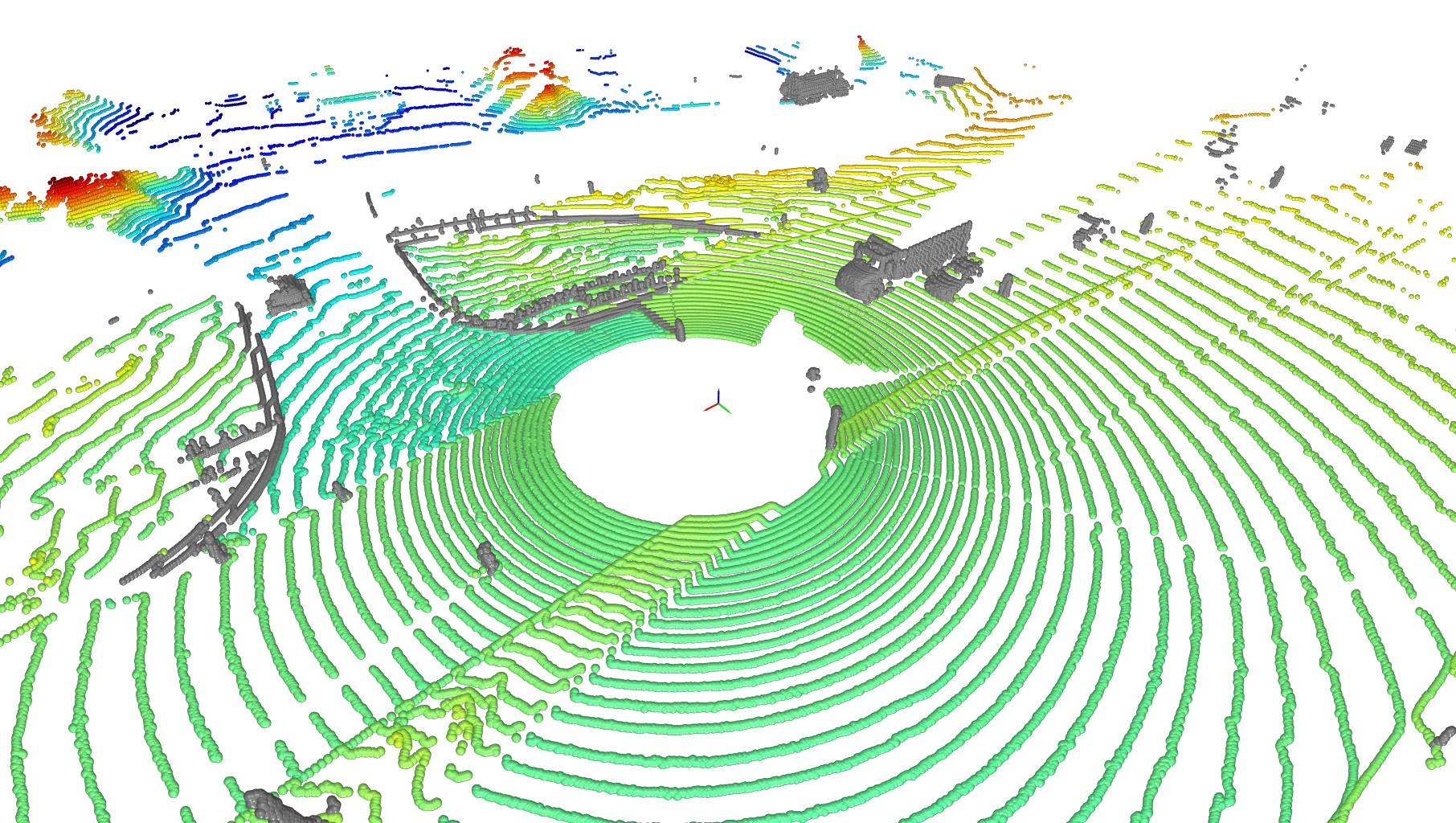}\\
    \vspace{2mm}
    \includegraphics[width=0.32\linewidth]{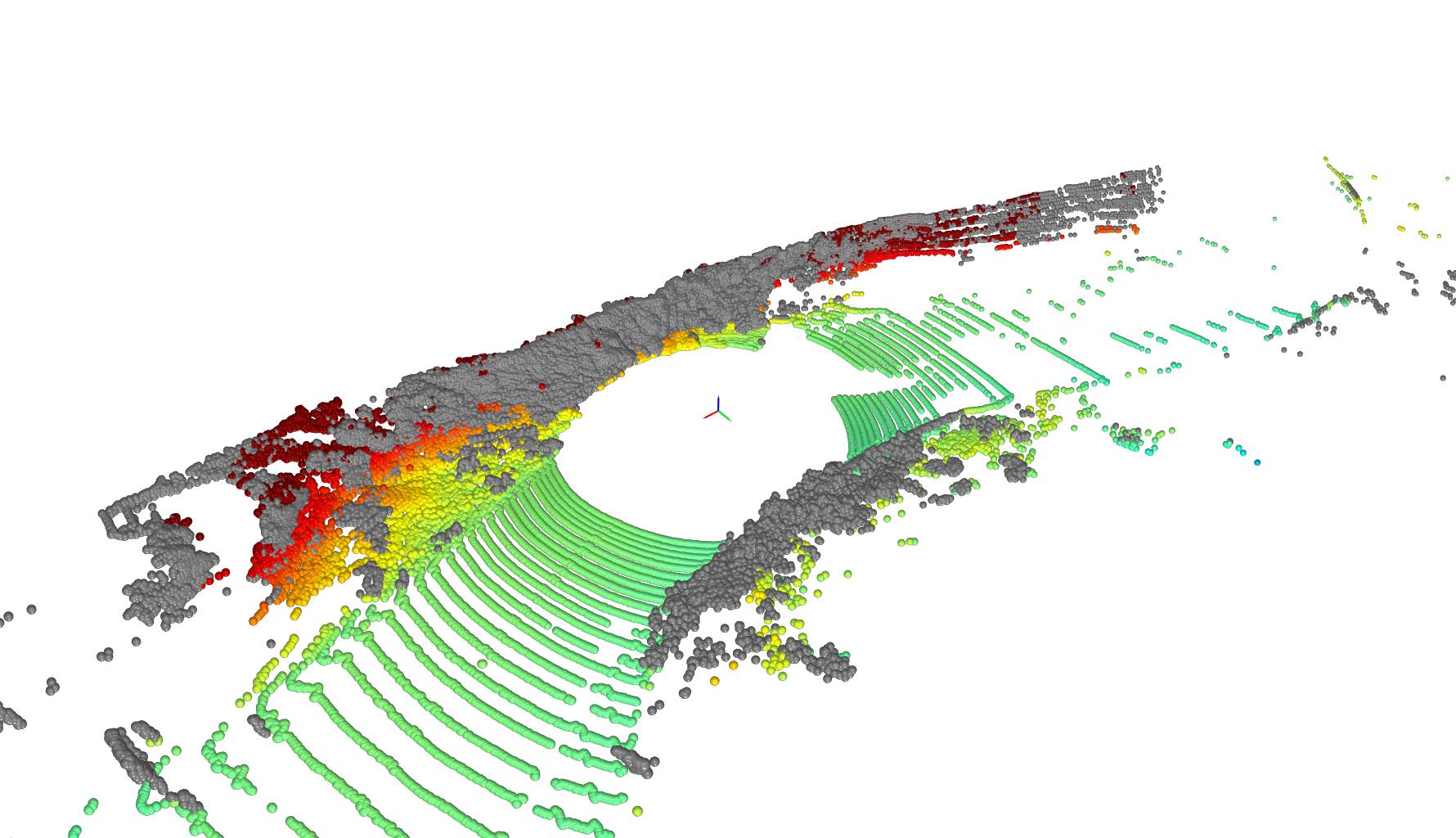}
    \includegraphics[width=0.32\linewidth]{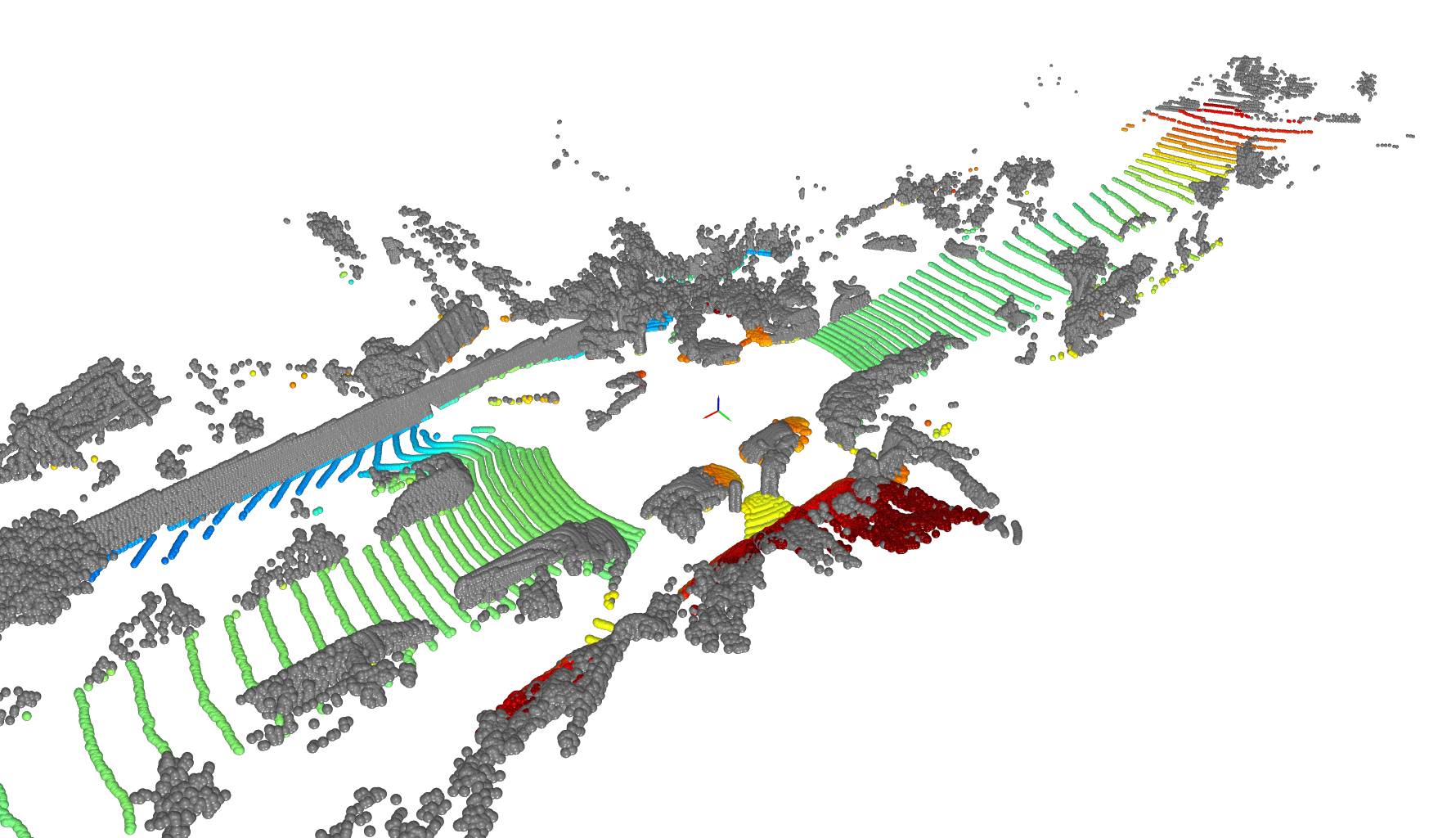}
    \includegraphics[width=0.32\linewidth]{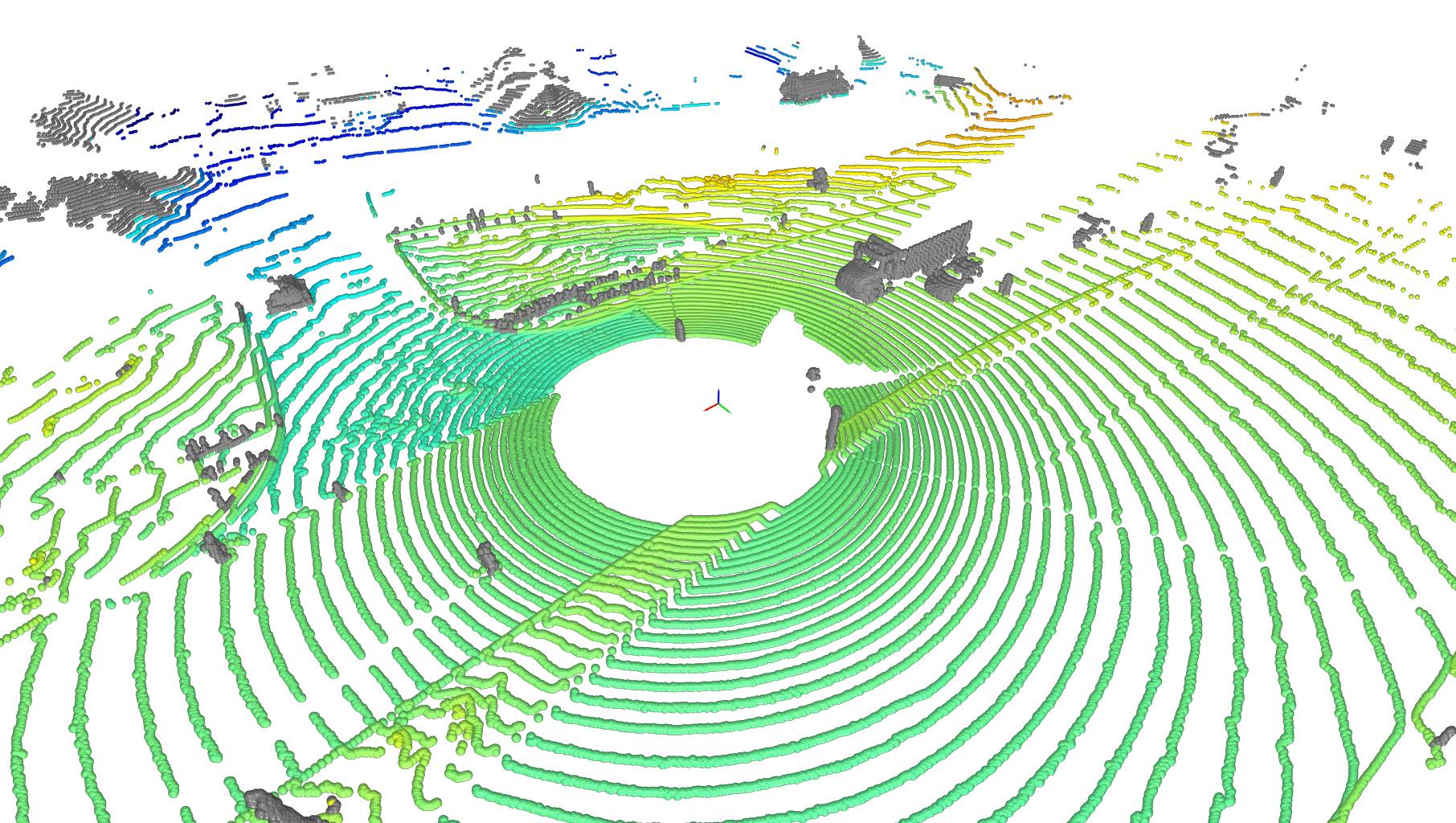}
    \caption{
    \textbf{Qualitative failure cases on Waymo Perception dataset~\cite{sun2020scalability}.} The top row shows ground truth labels, while the bottom row displays TerraSeg predictions. Ground points are color-coded by elevation; non-ground points are rendered in gray. From left to right, columns correspond to scan IDs \num{2513}, \num{4755}, and \num{2024}.}
    \label{fig:supp_fail_waymo}
\end{figure*}

While TerraSeg exhibits strong generalization capabilities, it occasionally encounters challenges in highly ambiguous geometric contexts.
Figures \ref{fig:supp_fail_nuscenes}, \ref{fig:supp_fail_semantickitti}, and \ref{fig:supp_fail_waymo} highlight specific failure cases across the three primary benchmarks.
As with the success cases, the top row provides the ground truth, and the bottom row illustrates the model's predictions.
The typical failure modes are the over-segmentation of dense, low-lying vegetation and atypical road debris that closely mimic the continuous geometric profile of the ground surface.

\begin{table*}[h]
    \centering
    \caption{
    \textbf{Hardware diversity in OmniLiDAR.} 
    A detailed breakdown of the \num{15} unique LiDAR sensor models utilized across the datasets.
    The datasets cover a wide spectrum of sensor configurations, including classic spinning LiDARs, solid-state variants, and FMCW sensors, ensuring robust cross-sensor generalization during training.
    }
    \label{tab:unique_lidars_appendix}
    \begin{tabularx}{\textwidth}{
    >{\raggedright\arraybackslash}p{3.8cm} | 
    >{\centering\arraybackslash}p{2.2cm} | 
    >{\centering\arraybackslash}p{1.4cm} | 
    >{\raggedright\arraybackslash}p{3.5cm} | 
    >{\raggedright\arraybackslash}X
    }
        \toprule
        \multicolumn{1}{l}{Sensor Model} & \multicolumn{1}{c}{Configuration} & \multicolumn{1}{c}{Rate} & \multicolumn{1}{l}{Dataset(s)} & \multicolumn{1}{l}{Description} \\
        \midrule
        Velodyne HDL-64E & 64 channels & \qty{10}{\hertz} & SemanticKITTI~\cite{behley2019semantickitti}, KITTI-360~\cite{liao2022kitti} & Classic high-resolution spinning LiDAR with full 360$^\circ$ FoV. \\
        Velodyne HDL-64 S3 & 64 channels & \qty{10}{\hertz} & View-of-Delft~\cite{palffy2022multi} & Newer HDL-64 generation with improved optics and timing. \\
        Velodyne HDL-32E & 32 channels & \qty{20}{\hertz} & nuScenes~\cite{caesar2020nuscenes} & Compact spinning LiDAR with higher temporal resolution. \\
        Velodyne VLP-32C & 32 channels & \qty{10}{\hertz} & Argoverse 2 Lidar~\cite{wilson2023argoverse} & Automotive-grade spinning LiDAR; dual-mounted in Argoverse 2. \\
        Velodyne VLS-128 & 128 channels & \qty{10}{\hertz} & Zenseact Open~\cite{alibeigi2023zenseact} & High-density long-range automotive spinning LiDAR. \\
        Velodyne VLP-16 & 16 channels & \qty{10}{\hertz} & Zenseact Open~\cite{alibeigi2023zenseact} & Short-range peripheral LiDARs for side and rear coverage. \\
        \midrule
        Waymo Mid-Range (TOP) & 64 channels & \qty{10}{\hertz} & Waymo Perception~\cite{sun2020scalability} & Roof-mounted panoramic LiDAR developed by Waymo. \\
        Waymo Short-Range & 4-5 channels & \qty{10}{\hertz} & Waymo Perception~\cite{sun2020scalability} & Near-field LiDARs (front, sides, rear) for close-range perception. \\
        \midrule
        Hesai Pandar64 & 64 channels & \qty{10}{\hertz} & PandaSet~\cite{xiao2021pandaset}, TruckScenes~\cite{fent2024man} & Long-range automotive-grade spinning LiDAR. \\
        Hesai PandarGT & 64 channels & \qty{10}{\hertz} & PandaSet~\cite{xiao2021pandaset} & Research-grade Pandar variant with higher angular accuracy. \\
        \midrule
        Ouster OS0 Rev 7 & 64 channels & \qty{10}{\hertz} & TruckScenes~\cite{fent2024man} & Wide-vertical-FoV LiDAR optimized for short-range perception. \\
        \midrule
        Aeva Wide-FoV (FMCW) & 110$^\circ$ FoV & \qty{10}{\hertz} & AevaScenes~\cite{narasimhan2025aevascenes} & FMCW LiDAR with wide angular coverage and velocity sensing. \\
        Aeva Narrow-FoV (FMCW) & 35$^\circ$ FoV & \qty{10}{\hertz} & AevaScenes~\cite{narasimhan2025aevascenes} & FMCW LiDAR optimized for long-range, high-precision sensing. \\
        \midrule
        ONCE Proprietary & 40 channels & \qty{10}{\hertz} & ONCE~\cite{mao2021one} & Medium-resolution automotive spinning LiDAR. \\
        Lyft Proprietary & Unknown & Unknown & Lyft Level 5~\cite{kesten2019lyft} & Undocumented hardware; conservatively counted as one unique model. \\
        \bottomrule
    \end{tabularx}
\end{table*}

\end{document}